\documentclass[11pt]{article}

\usepackage[margin=1in]{geometry}
\usepackage{microtype}
\usepackage[T1]{fontenc}
\usepackage{lmodern}

\usepackage{amsmath, amssymb, amsfonts}
\usepackage{bm}
\usepackage{gensymb}

\usepackage{graphicx}
\usepackage{caption}
\usepackage{subcaption}
\DeclareCaptionFont{mysize}{\fontsize{10}{10}\selectfont}
\usepackage{subcaption}
\usepackage{booktabs}
\usepackage{siunitx}

\usepackage{adjustbox}

\usepackage[colorlinks=true,allcolors=blue]{hyperref}

\usepackage{authblk}

\usepackage[numbers,sort&compress]{natbib}

\usepackage[nameinlink]{cleveref}

\begin{document}

\begin{center}
{\LARGE\bfseries
Enhanced Portable Ultra Low-Field Diffusion Tensor Imaging with Bayesian Artifact Correction and Deep Learning-Based Super-Resolution
\par}
\vspace{0.8em}
\end{center}

\begin{center}
{\normalsize
Mark D. Olchanyi\textsuperscript{1,2,3,4,*,$\dagger$},
Annabel Sorby-Adams\textsuperscript{5,6,*},
John Kirsch\textsuperscript{6},
Brian L. Edlow\textsuperscript{3,4,5,6},
Ava Farnan\textsuperscript{5},
Renfei Liu\textsuperscript{3,5},
Matthew S. Rosen\textsuperscript{6},
Emery N. Brown\textsuperscript{1,2,4,6},
W. Taylor Kimberly\textsuperscript{3,5,6,**},
Juan Eugenio Iglesias\textsuperscript{6,7,8,**},
for the Alzheimer’s Disease Neuroimaging Initiative\textsuperscript{$\ddagger$}
\par}
\end{center}

\vspace{0.8em}

{\footnotesize
\noindent \textsuperscript{1}Picower Institute, Massachusetts Institute of Technology, Cambridge, MA, USA, 02139\\
\textsuperscript{2}Institute for Medical Engineering and Science, Massachusetts Institute of Technology, Cambridge, MA, USA, 02142\\
\textsuperscript{3}Center for Neurotechnology and Neurorecovery, Massachusetts General Hospital and Harvard Medical School, Boston, MA, USA, 02114\\
\textsuperscript{4}MIT/MGH Brain Arousal State Control Innovation Center, Massachusetts Institute of Technology, Cambridge, MA, USA, 02139\\
\textsuperscript{5}Department of Neurology, Massachusetts General Hospital and Harvard Medical School, Boston, MA, USA, 02114\\
\textsuperscript{6}Athinoula A. Martinos Center for Biomedical Imaging, Massachusetts General Hospital and Harvard Medical School, Charlestown, MA, USA, 02129\\
\textsuperscript{7}Computer Science and Artificial Intelligence Laboratory, Massachusetts Institute of Technology, Cambridge, MA, USA, 02139\\
\textsuperscript{8}Hawkes Institute, University College London, London, UK, WC1E 6BT
\par}

\vspace{0.8em}

\noindent{\footnotesize
\textsuperscript{*}co-first authors\\
\textsuperscript{**}co-senior authors\\
\textsuperscript{$\dagger$}Corresponding author: \href{mailto:olchanyi@mit.edu}{olchanyi@mit.edu}\\
\textsuperscript{$\ddagger$}Data used in preparation of this article were obtained from the Alzheimer’s Disease Neuroimaging Initiative (ADNI) database (adni.loni.usc.edu). As such, the investigators within the ADNI contributed to the design and implementation of ADNI and/or provided data but did not participate in analysis or writing of this report. A complete listing of ADNI investigators can be found on the \href{https://adni.loni.usc.edu}{ADNI website}.
}

\vspace{0.7em}

\begin{abstract}
Portable, ultra-low-field (ULF) magnetic resonance imaging has the potential to expand access to neuroimaging but currently suffers from coarse spatial and angular resolutions and low signal-to-noise ratios. Diffusion tensor imaging (DTI), a sequence tailored to detect and reconstruct white matter tracts within the brain, is particularly prone to such imaging degradation due to inherent sequence design coupled with prolonged scan times. In addition, ULF DTI scans exhibit artifacting that spans both the space and angular domains, requiring a custom modelling algorithm for subsequent correction. We introduce a nine-direction, single-shell ULF DTI sequence, as well as a companion Bayesian bias field correction algorithm that possesses angular dependence and convolutional neural network-based superresolution algorithm that is generalizable across DTI datasets and does not require re-training (“DiffSR”). We show through a synthetic downsampling experiment and white matter assessment in real, matched ULF and high-field DTI scans that these algorithms can recover microstructural and volumetric white matter information at ULF. We also show that DiffSR can be directly applied to white matter-based Alzheimer’s disease classification in synthetically degraded scans, with notable improvements in agreement between DTI metrics, as compared to un-degraded scans. We freely disseminate the Bayesian bias correction algorithm and DiffSR with the goal of furthering progress on both ULF reconstruction methods and general DTI sequence harmonization. We release all code related to DiffSR for public use\textsuperscript{\S}.

\end{abstract}

\vspace{1.5em}

\begingroup
\renewcommand{\thefootnote}{\S}
\footnotetext{Code is available at: \href{https://github.com/markolchanyi/DiffSR}{https://github.com/markolchanyi/DiffSR}.}
\endgroup

\section{Introduction}
Magnetic resonance imaging (MRI) is an indispensable neuroimaging method for both research and clinical brain assessment, yet its utility is often constrained by long acquisition times and scanning cost, as well as the need to dedicated imaging suites and personnel. This significantly hinders access to MRI, especially in rural settings and patient populations where transport to scanners is infeasible. The recent emergence of portable MRI systems which operate at ultra-low magnetic field (ULF) strengths now makes it possible to perform point-of-care MR-based imaging in a cost-effective manner in settings without access to conventional high-field (HF) MRI scanners \citep{Liu2021,Sheth2021,Cooley2020}. ULF MRI systems accomplish this with permanent magnets to generate static magnetic fields at <100 milli-Tesla (mT) field strengths, with gradient and radiofrequency coil systems integrated in simpler head coils, all operating at standard wall power. The absence of cryogenically cooled coils simplifies deployment but constrains acquisitions in terms of signal-to-noise ratio (SNR), resolution, and scan time due to the same lack of active cooling. These limitations are amplified in the application of ULF to advanced MRI sequences such as diffusion MRI (dMRI), which is currently the only non-invasive imaging modality that can map and probe white matter microstructure within the human brain. Specifically, dMRI applies direction-specific sampling of proton diffusion through transverse magnetic field gradients \citep{Stejskal1965}, allowing for accurate voxel-wise measurement of white matter integrity and orientation profiles \citep{Tuch2002,Tournier2004}. As such, dMRI has proved useful for many research and clinical applications, including acute stroke monitoring, white matter assessment in neurodegenerative disorders \citep{Nir2013,Bozzali2002,Zhang2020PD,Filippi2019,Winter2021}, tractography of white matter disconnections in traumatic brain injury \citep{Edlow2013,Kinnunen2011,Bourke2021,Olchanyi2026}, and neuro-navigation for electrode placement for deep brain stimulation \citep{Calabrese2016}. Incorporating dMRI to ULF is a particularly attractive application for outpatient neurodegenerative disease clinics, critical care settings and surgical suites due to its portability and ease-of-use.

\vspace{0.5cm}

Among dMRI techniques, diffusion tensor imaging (DTI) is the most simplistic, and thus most feasible for use at ULF, due to its estimation of diffusion properties within brain tissue from minimal numbers of diffusion-encoding gradients \citep{LeBihan2001}. For simplicity of notation, we will hereafter refer to all mentioned dMRI sequences as “\textit{DTI}”. Early ULF DTI methods show its feasibility \citep{Gholam2025}, but also reveal several of its failure modes. These include low SNR, high partial voluming due to large voxel size, broad point spread functions due to fast spin echo readout duration, spatially dependent diffusion encoding nonuniformity from B0 gradients, and significant diffusion-sensitized signal intensity variation in both the spatial and radial domain. What separates the latter artifact is its uniqueness to ULF, as directional-dependence is not assumed in any standard DTI bias field correction algorithms. Hardware-level correction of these artifacts is possible but limited in scope due to the underlying properties of ULF permanent magnets, coil configurations, and cooling-system designs. As such, we attempt to correct for B0 gradient-based and diffusion-sensitized signal variations \textit{post hoc} by approximating them as smooth, multiplicative, directionally dependent bias fields. We optimize bias field parameters through Bayesian inference, over a likelihood function conditioned on distributions of fractional anisotropy and eigenvector directionality, the priors for which are generated with a probabilistic atlas from HF DTI. This approach allows us to capture smooth variations in direction-dependent intensity while preserving legitimate high-FA white matter contrast. Because this model optimizes over peak directionality and is regularized by microstructural properties, it is robust to sequence variations and exact number of diffusion-encoding gradients, making it broadly applicable across ULF protocols. 

\vspace{0.5cm}

Even after standard preprocessing and model-based correction of smooth intensity variations, residual artifacts exacerbated at ULF such as shot-to-shot instability, eddy currents and thermal/motion-based noise remain, all at poor spatial and angular resolutions of ULF DTI. Taken together, this significantly narrows the scope of analysis at ULF versus HF. Spatial and angular superresolution ULF DTI data would greatly aid in narrowing this gap and would provide a method to for ULF DTI sequence harmonization. Superresolution of DTI data can be formatted as either spatial, angular, or joint spatio-angular superresolution tasks. Spatial superresolution maintains the original angular sampling scheme and spatially enhances (in either image space or k-space) each diffusion-encoded channel separately. Spatial DTI superresolution has historically been accomplished through concatenating sets of 3-D diffusion gradient volumes with angular context, usually with a paired set of b-vector tokens and training them to match image-token pairs with convolutional encoder models \citep{Lyon2022}.  Pure angular superresolution aims to solely enhance sampling in the angular (Q-space) domain. Such a task can either result in a set sample scheme, such as a static set of target b-vectors, or a continuous Q-space representation, allowing for a target b-vector set of arbitrary size, angle, or even magnitude (which also results in an arbitrary b-value shell magnitude). Prior algorithms have mostly accomplished through deep-learning methods to match the quality of training-set q-space samples with discrete representations \citep{Ordinola2025,Chen2022,Chen2020}, projections into a continuous spherical space parameterized by basis functions \citep{Rathi2014,Nath2020} such as fiber orientation distribution functions (ODF) \citep{Zeng2022}, or continuous nonparametric representations \citep{Ewert2024}. To date, all superresolution techniques are specifically tailored to operate on conventional HF DTI data. We therefore propose a joint spatio-angular superresolution method that is optimized on training data that matches the spatial resolution, angular resolution, SNR, and degradation patterns of ULF DTI, which we term DiffSR. DiffSR processes ULF DTI data that has been transformed into a spherical harmonic (SH) representation, which provides a input size invariance. Prior studies have used SH representations of DTI data for sequence harmonization \citep{DeLuca2022}, deep learning-based denoising algorithms \citep{Chen2025} and representation of fiber orientation distributions 
\citep{Tournier2019}. DiffSR is built on a U-Net convolutional neural network (CNN) model flanked by multilayer perceptron (MLP)-based transformers operating on icosahedral projections of SH coefficients. Critically, we tailor DiffSR to ULF contrasts and resolutions by aggressive augmentation of Human Connectome Project (HCP) Young Adult training data directly in SH space.

\vspace{0.5cm}

In this work, we introduce a practical 9-direction ULF DTI sequence and companion software tailored to its constraints and degradation patterns: a Bayesian bias field correction algorithm that directly incorporates angular information of the diffusion signal and DiffSR, a joint spatio-angular superresolution method that operates in SH space. We test the robustness of DiffSR through a synthetic spatial and angular downsampling experiment. We then show that Bayesian bias field correction coupled with DiffSR accurately recovers white matter-specific information when applied to an 18-subject cohort with matched ULF and HF DTI scans. Finally, we show the clinical translatability of DiffSR in a volumetric classification task on Alzheimer's disease (AD) data. Through these algorithms, we enable portable, ULF DTI to be more clinically viable and provide a method for ULF DTI harmonization across scanning sites. We disseminate all code for the correction and superresolution algorithms through \href{https://github.com/markolchanyi/DiffSR}{GitHub}. 

\section{Materials and Methods}
\subsection{Ultra-low field and matched high field DTI acquisition information}
18 Participants (7 Females; 11 Males: 32.1$\pm$12.9 years) underwent LF DTI scanning on a 64mT Hyperfine Swoop scanner with a permanent magnet system (Hyperfine Inc., hardware Mk 1.9, software 8.8.0). Subject-specific information for each participant can be found in Supplementary Table $\ref{tab:suppT1}$. Our DTI sequence was a three-dimensional multi-shot diffusion-weighted fast spin echo (MS-DWFSE) single-shell sequence (TR=700ms, TE=77.7ms, flip angle=90°), with nine diffusion-encoding gradient directions (b=700$\frac{s}{mm^2}$), and three low-b volumes (b=0$\frac{s}{mm^2}$) interlaced between directions 2 and 3, directions 5 and 6, and directions 8 and 9 as to facilitate passive scanner cooling. Whole brain scans were performed with a field of view of 56 by 64 by 52 voxels, at a 3.5mm isotropic spatial resolution. The total DTI sequence duration was 58 minutes. For each participant, a matched HF DTI scan was performed in a 3T Siemens Prisma scanner (Siemens Healthineers, Erlangen, Germany) with a 2-dimensional single-shot echo planar imaging (EPI) sequence (TR=4000ms, TE=60ms, flip angle=90°) with anterior-posterior phase encoding, 64 diffusion-encoding gradient directions (b=900$\frac{s}{mm^2}$) and 9 low-b volumes at a 1.8mm isotropic spatial resolution. This study and protocol was approved by the Mass General Brigham Institutional Review Board (protocol number: 2022P001892).

\subsection{External DTI datasets}
DiffSR is trained on 100 subjects from the WU-Minn HCP Young Adult dataset \citep{VanEssen2013} (Sections 2.5 and 2.6). All WU-Minn HCP subjects were scanned on a 3T Siemens Skyra Connectom scanner (Siemens Healthineers, Erlangen, Germany) with a spin-echo EPI DTI sequence (TR=5520ms, TE=89.5ms, flip angle=78$\degree$) with 90 diffusion encoding gradient directions acquired at three separate shell values (b=1000$\frac{s}{mm^2}$, b=2000$\frac{s}{mm^2}$, b=3000$\frac{s}{mm^2}$) and six low-b volumes at a 1.25mm isotropic spatial resolution. We henceforth call this dataset the “\textit{WU-Minn HCP dataset}”. 

\vspace{0.5cm}

To directly quantify how DiffSR restores diffusion contrast and microstructural information from synthetically degraded (Section 3.1) in vivo DTI data, we used 30 subjects from the MGH Adult HCP dataset \citep{Setsompop2013}. Each MGH HCP Adult subject was scanned in a custom 3T Siemens Connectom scanner (Siemens Healthineers, Erlangen, Germany) with a Spin echo EPI sequence (TR=8800ms, TE=57ms) with 64 diffusion encoding gradient directions interspersed across two shells (b=1000$\frac{s}{mm^2}$ and b=3000$\frac{s}{mm^2}$) and 128 diffusion encoding gradient directions across two additional shells (b=5000$\frac{s}{mm^2}$ and b=10000$\frac{s}{mm^2}$) and 40 interspersed low-b volumes at 1.5mm isotropic spatial resolution. We henceforth call this dataset the “\textit{Connectom HCP dataset}”. 

\vspace{0.5cm}

Data used in the preparation of this article were obtained from the Alzheimer’s Disease Neuroimaging Initiative (ADNI) database (\href{adni.loni.usc.edu}{adni.loni.usc.edu}). The ADNI was launched in 2003 as a public-private partnership, led by Principal Investigator Michael W. Weiner, MD. The primary goal of ADNI has been to test whether serial MRI, positron emission tomography (PET), other biological markers, and clinical and neuropsychological assessment can be combined to measure the progression of mild cognitive impairment (MCI) and early AD. 

\vspace{0.5cm}

To test whether spatio-angular superresolution can enhance the analysis of individual white matter tracts associated with AD/MCI (Section 3.2), we applied DiffSR to 24 AD subjects (10 Females; 14 Males: 73.7±9.0 years), 15 subjects with late-stage MCI (LMCI) (9 Females; 6 Males: 74.6±7.7 years), and 138 control subjects (76 Females; 62 Males: 74.0±7.6 years) from the ADNI database \citep{Petersen2010,Weiner2017}. For this analysis, we grouped the AD and LMCI groups into one AD/LMCI supergroup (19 Females; 20 Males: 74.0±8.4 years). All subjects were scanned with a single-shell ADNI3-Basic axial sequence with TR=7200ms, TE=56ms, flip angle=90° at a b-value 1000$\frac{s}{mm^2}$ with 48 diffusion-encoding directions and 7 b=0 images acquired at 2mm isotropic spatial resolution. One control subject was excluded from analysis due to preprocessing failure.

\subsection{DTI data preprocessing}

All DTI sequences used in this study underwent standard preprocessing using MRtrix \citep{Tournier2019} and FSL \citep{Jenkinson2012,Smith2004} software packages. This included (1) denoising with the MRtrix \textit{dwidenoise} command, (2) Gibbs de-ringing with the MRtrix \textit{mrdegibbs} command, (3) motion and eddy-current distortion correction with FSL and incorporated in the MRtrix \textit{dwifslpreproc} command, and (4) stationary bias field correction with the N4 ANTs algorithm incorporated in the MRtrix \textit{dwibiascorrect} command. For ULF DTI preprocessing, we did not perform stationary bias field correction (step 4 above) due to the presence of direction-specific bias fields which we explicitly correct for with a custom algorithm (see section 2.4). In addition to standard preprocessing, we applied ULF-specific within-subject motion correction, which is relatively more pronounced due to the overall scan time and increase freedom to head motion within the ULF head coil, via explicit co-registration. Specifically, we generated synthetic T1-weighted images for all low-b and diffusion encoding gradient direction volumes via the FreeSurfer SynthSR algorithm \citep{Iglesias2023} to harmonize low-b and diffusion encoding gradient direction contrast and localize subcortical regions of interest (ROIs) including the left/right thalamic, caudate, putamen, and lateral ventricle subfields using the FreeSurfer Supersynth algorithm \citep{Liu2025}. We then co-registered the second and third low-b volumes and all nine diffusion-weighted volumes to the first low-b with rigid followed by affine followed by SyN registration with each subcortical ROI used as a fiducial with Demons-based loss using the ANTs software package \citep{Avants2011}. All b-vector tables were rotated in accordance with the calculated affine transform matrices.

\vspace{0.5cm}

For tract-wise validation analyses in sections 3.2 and 3.3, we generated all white matter ROIs in HF DTI space with the \textit{Tracula} algorithm \citep{Yendiki2011,Maffei2021}, for which as a prerequisite we processed each subject with FreeSurfer’s recon-all-clinical software pipeline \citep{Gopinath2025} on the mean low-b volume. For ULF DTI analysis in section 3.3, all \textit{Tracula} white matter ROIs were affinely co-registered to ULF space. For deterministic tractography comparisons, as discussed in section 3.3 and visualized in Figure 8, we generated all deterministic streamlines with the MRtrix tckgen command using the iFOD2 algorithm. For each \textit{Tracula} white matter tract segmentation mask, deterministic streamlines were propagated through 2-4 evenly-spaced segmentation mask cross-sections, which were visually inspected and manually corrected for in ULF space to account for deviations after affine registration from HF space. Streamline exclusion regions were also defined for a subset of tract ROIs, such as the posterior aspect of the brain for the anterior thalamic radiation, or the arcuate fasciculus ROI for the superior longitudinal fasciculus streamlines.

\subsection{Bias field estimation and correction for LF DTI}

The $i$'th diffusion-weighted signal $S_i$ along a diffusion-encoding gradient direction $\mathbf{u}_i$ in a voxel $x \in \Omega$ (where $\Omega$ is image space) can be modeled by the Stejskal-Tanner equation \citep{Stejskal1965}: 

\begin{equation}
S_i(x) = S_0(x)\exp\!\left[-\,b_i\,\mathbf{u}_i^{\mathsf T}\,\mathbf{D}(x)\,\mathbf{u}_i\right],
\end{equation}

\noindent where $\mathbf{D}(x) \in {{\mathbb{S}}_+^3}$ is the full, symmetric diffusion tensor, $S_0$ is the non-diffusion-weighted (i.e., low-b) signal, and $b_i$ is the b-value. Empirically, this signal is degraded by several artifacts during ULF-DTI acquisitions, including: (1) B1 bias fields, which can be modeled by a low-frequency multiplicative field and are stationary across gradient directions $\{\mathbf{u}_i \}$; (2) B0 inhomogeneities, which can be modeled as a spatially smooth linear scaling to the b-values $\{b_i \}$ and are also directionally-independent; and (3) further spatially smooth intensity modulations that vary across diffusion-encoding gradients, possibly due to eddy currents, concomitant fields/Maxwell terms, and/or motion artifacts \citep{Gholam2025}.

\vspace{0.5cm}

\begin{figure}[t]
    \centering
    \includegraphics[width=1.0\linewidth]{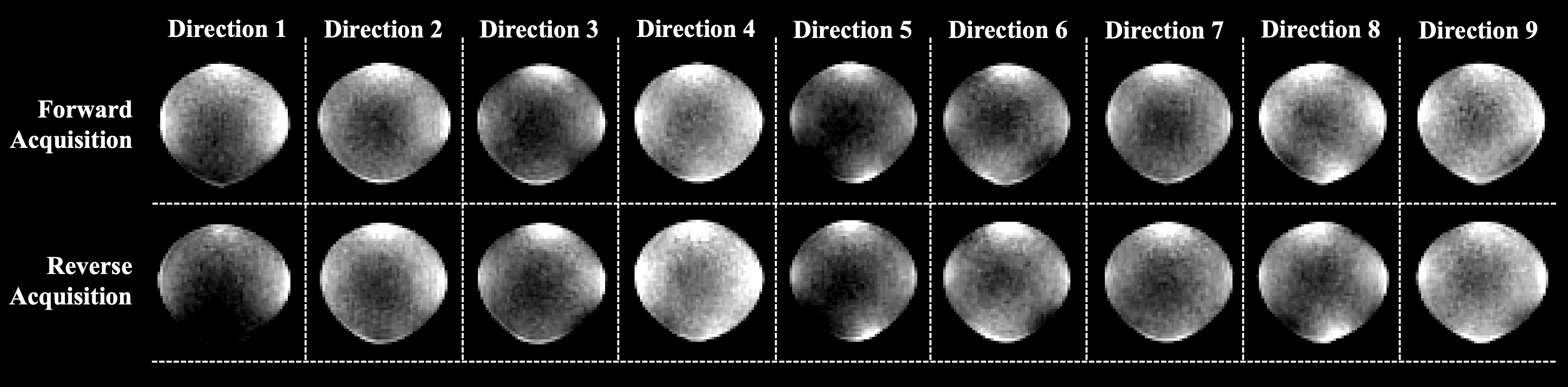}
    \caption{\textbf{Direction-specific bias fields encountered in ULF DTI}. Application of our nine-direction ULF DTI sequence to a spherical 40$\%$ Polyvinylpyrrolidone phantom displays smooth intensity fields that vary with diffusion encoding gradient direction. The top row displays the forward (i.e., standard) ULF DTI acquisition with diffusion weighting (excluding the interleaved low-b acquisitions), while the bottom row displays the same ULF DTI sequence but with diffusion encoding gradient directions acquired in reverse order.}
    \label{fig:Figure_1}
\end{figure}

Previous work on ULF DTI \citep{Gholam2025} has attempted to mitigate these artifacts with a combination hardware-level and postprocessing techniques, including B1 power calibration, standard bias field estimation with the b0 image, and B0 shimming. However, these methods do not address smooth biases that vary across diffusion-encoding gradient directions. Representative, direction-dependent ULF-DTI bias fields are illustrated in Figure $\ref{fig:Figure_1}$, which show reconstructions from different diffusion-encoding gradients applied to a 40$\%$ Polyvinylpyrrolidone (PVP) phantom (National Institutes of Health: CNRH). During the forward (i.e., standard) ULF-DTI acquisition, bias fields are noticeably both spatially and directionally varying. However, no temporal and/or coil temperature variations of the bias fields are observed, as confirmed by acquiring ULF-DTI gradient volumes in reverse order (i.e., direction 9 acquired first and direction 1 acquired last, bottom row), indicating that the induced bias depends on the applied gradient rather than being dominated by time-varying effects.

\vspace{0.5cm}

To explicitly correct for directionally varying bias fields unique to ULF-DTI, we follow a purely computational approach. Specifically, we model these biases as smooth, multiplicative fields for the signal and b-value, leading to the modified Stejskal-Tanner equation:

\begin{equation}
S_i(x)=\Gamma_i(x)\,S_0(x)\,\exp\!\left[-\Upsilon_i(x)\,b_i\,\mathbf{u}_i^{\mathsf T}\,\mathbf{D}(x)\,\mathbf{u}_i\right],
\end{equation}

\noindent where $\Gamma_i(x)$ is a direction-dependent signal bias with an additional b-value modulated (e.g., Maxwell terms) component $\Upsilon_i(x)$. We perform all correction in the log-domain: 

\begin{equation}
y_i(x) \equiv \log S_i(x)
= \log S_0(x) - \Upsilon_i(x)\, b_i\, \mathbf{u}_i^{\mathsf T}\mathbf{D}(x)\mathbf{u}_i
+ \log \Gamma_i(x)
= \log S_0(x) + \zeta_i(x).
\end{equation}

For simplicity, we collapse the signal and b-value modulated bias terms from Eq. (3) into a single smooth log-bias that also implicitly depends on $\mathbf{D}(x)$:

\begin{equation}
\zeta_i(x) = \log \Gamma_i(x) - \Upsilon_i(x)\, b_i\, \mathbf{u}_i^{\mathsf T}\mathbf{D}(x)\mathbf{u}_i
\end{equation}

Following previous work in the Bayesian segmentation and bias field estimation literature \citep{VanLeemput1999}, we model $\zeta_i(x)$  as a linear combination of $N$ smooth basis functions $\{\Phi_n(x) \}$:  specifically, as a sum of discrete cosine transform (DCT) functions:

\begin{equation}
\zeta_i(x) = \sum_{n=1}^{N} c^{\zeta}_{i,n}\,\Phi_n(x),
\end{equation}

\noindent where we set N=6. Therefore, our problem amounts to estimating the sets of coefficients $c^{\zeta}_{i,n}$. We pose this as a Bayesian inference problem within a generative model that uses a simple parametric prior: a linearly registered atlas of fractional anisotropy (FA) and principal eigenvector $\mathbf{v1}$. We parameterize these priors with a combination of a tissue class-conditioned Beta distribution for FA and a Dimroth-Scheidegger-Watson (DSW) distribution for $\mathbf{v1}$, as used in prior DTI modelling and segmentation tasks \citep{Iglesias2019,Tregidgo2023BAYES}:

\begin{equation}
\begin{aligned}
p\!\left(\mathrm{FA}(x)\mid L(x)=t\right) &\sim \mathrm{Beta}\!\left(\alpha_{L(x)}(x),\,\beta_{L(x)}(x)\right),
\qquad t \in \{\mathrm{WM},\mathrm{GM},\mathrm{CSF}\},\\[0.6em]
p\!\left(\mathbf{v1}(x)\right) &\propto \exp\!\left(\kappa(x)\,\mathrm{FA}(x)\,\bigl|\mathbf{v1}_{\mu}(x)^{\mathsf T}\mathbf{v1}(x)\bigr|^{2}\right),
\end{aligned}
\end{equation}

\noindent where $\alpha_{L(x)}(x)$ and $\beta_{L(x)}(x)$ are estimated via method-of-moments based on voxel-wise categorical assignment $L(x)$ to one of three tissue classes: white matter (WM), gray matter (GM) or cerebrospinal fluid (CSF). $\mathbf{v1}_{\mu}(x)$ is the directional DSW prior with the attenuation parameter $\kappa(x)$, which controls the “sharpness” of the prior peak around $\mathbf{v1}_{\mu}(x)$ and which we derive from the normalized principal eigenvalue of the atlas diffusion tensor $\tilde{\lambda}_1 = \frac{\lambda_1}{\lambda_1 + \lambda_2 + \lambda_3}$ (where all eigenvalues are clamped between $\frac{1}{3}$ and 1):

\vspace{-0.4cm}

\begin{equation}
\kappa(x) = \max\!\left\{\frac{3\,\tilde{\lambda}_1(x)-1}{1-\tilde{\lambda}_1(x)}\right\}.
\end{equation}

\vspace{0.2cm}

\begin{figure}[t]
    \centering
    \includegraphics[width=0.6\linewidth]{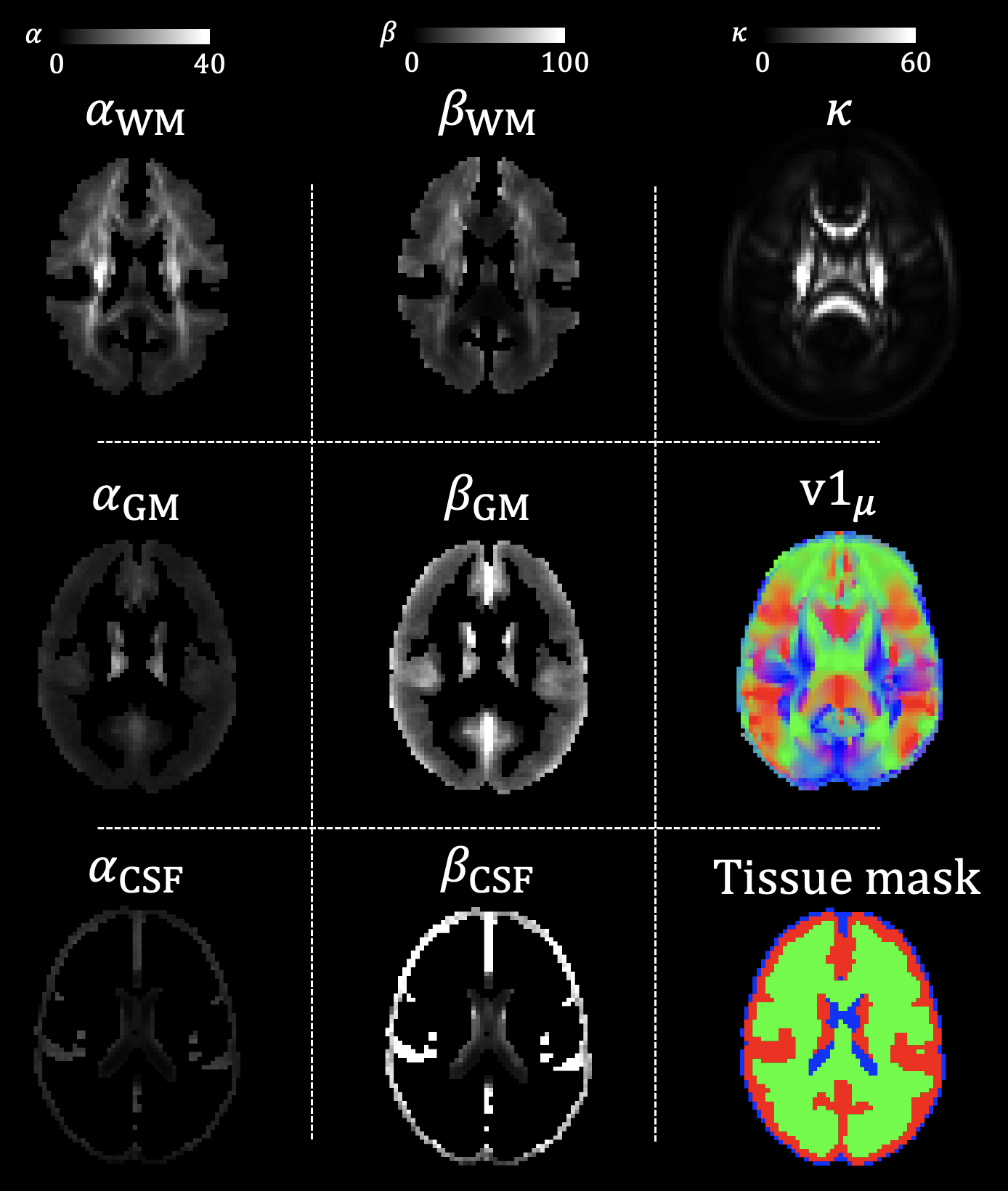}
    \caption{\textbf{Beta and DSW parametric priors for bias correction}. Shown in the left two columns are the HCP priors for beta distribution coefficients separated by tissue class: white matter, grey matter and cerebrospinal fluid. Shown in the right column are the directional ($\mathbf{v1}_{\mu}$) and attenuation ($\kappa$) priors for the DSW distribution, as well as the hard tissue segmentations in MNI space used for $\alpha_L$ and $\beta_L$ tissue class separation.}
    \label{fig:Figure_2}
\end{figure}

We generate all priors by affinely co-registering DTI scans from 860 subjects from the WU-Minn HCP dataset \citep{VanEssen2013,Sotiropoulos2013} to Montreal Neurological Institute (MNI) space and fitting the parameters of the two distributions at each voxel. The affine transform matrices are estimated with the ANTs software package \citep{Avants2011} using the b0 images and are used to deform the FA and V1 maps, rotating the latter with a tensor reorientation algorithm proposed by Zhang et. al. \citep{ZHANG2006}. The Beta and DSW atlas coefficients are illustrated in Figure $\ref{fig:Figure_2}$.

\vspace{0.2cm}

Given a subject DTI, one simply deforms the Beta and DSW priors from MNI space to subject space (rotating $\mathbf{v1}_{\mu}(x)$ of the DSW distribution via tensor reorientation) and numerically optimizes the log-likelihood of the observed data with respect to $c^{\zeta}_{i,n}$. To do this, we extract FA(x) and $\mathbf{v}_1(x)$ via a least-squares fit and eigen decomposition on $\mathbf{D}(x)$ at each optimization step and generate the negative log-likelihoods:

\begin{equation}
\begin{aligned}
\mathcal{L}\!\left(\alpha_{L(x)},\beta_{L(x)} \mid x; L(x)\right)
&= -\log\!\left(p[\mathrm{FA}(x)]\right) \\
&= -\Bigl[\bigl(\alpha_{L(x)}(x)-1\bigr)\log(\mathrm{FA}(x))
+ \bigl(\beta_{L(x)}(x)-1\bigr)\log\!\bigl(1-\mathrm{FA}(x)\bigr)\Bigr],\\[0.9em]
\mathcal{L}(\kappa \mid x)
&= -\log\!\left(p[\mathbf{v1}(x)]\right)
= -\kappa(x)\,\mathrm{FA}(x)\,\bigl[\mathbf{v1}_{\mu}(x)^{\mathsf T}\mathbf{v1}(x)\bigr]^2 .
\end{aligned}
\end{equation}

We determine optimal DCT bias coefficients $\widehat{\{c^{\zeta}_{i,n}\}}$ via maximum a posteriori estimation with regularization on coefficient magnitudes and FA values within gray matter tissue, formulated as:

\vspace{-0.2cm}

\begin{equation}
\widehat{\{c^{\zeta}_{i,n}\}} \;\propto\;
\arg\min_{\{c^{\zeta}_{i,n}\}}
\left\{
\sum_{x} \Bigl[-\log\!\bigl(p[\mathrm{FA}(x)]\bigr) - \log\!\bigl(p[\mathbf{v1}(x)]\bigr)\Bigr]
\;+\; \lambda_c \sum_{i,n} \bigl[c^{\zeta}_{i,n}\bigr]^2
\;+\; \lambda_{\mathrm{GM}} \sum_{x\in \mathrm{GM}} \bigl[\mathrm{FA}(x)\bigr]^2
\right\}.
\end{equation}

$\widehat{\{c^{\zeta}_{i,n}\}}$ is determined iteratively via backpropagation with Adaptive Moment Estimation (ADAM) burn-in \citep{Kingma2014} followed by Limited-memory Broyden-Fletcher-Goldfarb-Shanno (LBFGS) optimization; the former excels at jumping over local optima, whereas the latter ensures convergence to a local minimum. All the operations connecting the diffusion-weighted signals $\{S_i(x) \}$ and the FA/V1 maps are differentiable. Of note, the bias field model is identifiable only up to a common multiplicative factor applied uniformly to all gradient directions. To remove this indeterminacy, we correct for the bias field in the mean low-b (also modeled with DCT coefficients) a priori with a tissue-specific approach on soft tissue labels obtained through Supersynth, which we solve with Expectation Maximization. The low-b bias field is guaranteed to be spatially centered and consistent across tissues, yielding the bias-corrected $\widehat{S_0}$. We therefore divide $S_i$ by $S_0$ before direction-specific optimization, which removes the overall multiplicative ambiguity and allows for correct centering of the final direction-dependent bias field estimate.

\begin{figure}[t]
    \centering
    \includegraphics[width=0.99\linewidth]{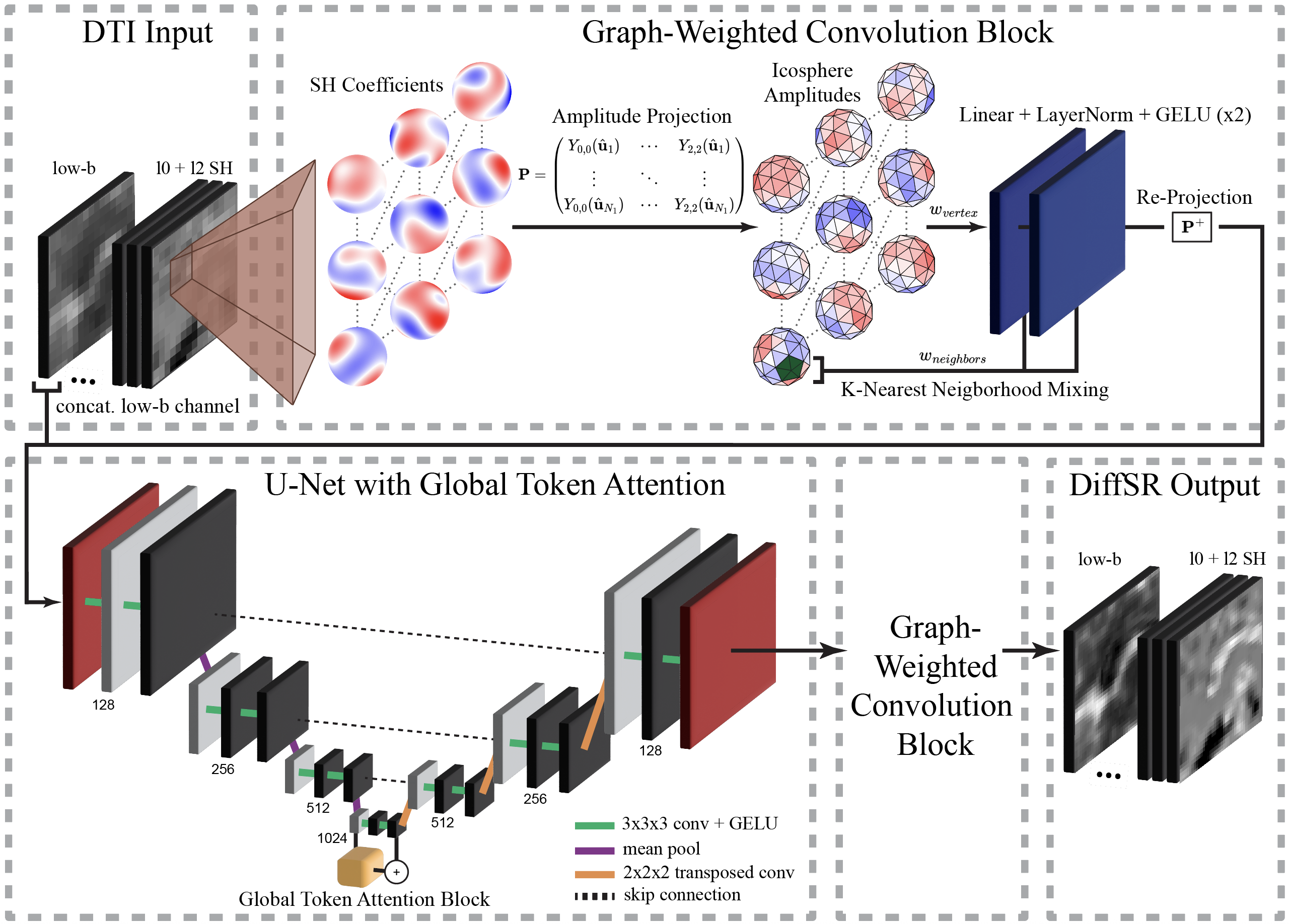}
    \caption{\textbf{Overview of the DiffSR forward pipeline}. The pipeline consists of an icosahedral projection of the spherical harmonic coefficients derived from the raw DTI signal, followed by two layers of graph-weighted convolutions to capture proximal angular dependence. The filtered projected amplitudes are then re-projected back to spherical harmonic space via matrix inversion, which we call a graph-weighted convolution block. The transformed spherical harmonic coefficients are then propagated through a U-Net CNN with an eight global token attention block at the bottleneck. Finally, the propagated SH coefficients are filtered through a second graph-weighted convolution block to generate the superresolved DTI volume.}
    \label{fig:Figure_3}
\end{figure}

\subsection{Augmentation strategies for neural network training}

For compactness and generalizability to ULF, we train DiffSR to perform superresolution directly on the second-order SH coefficient estimates of the DTI signal. DiffSR is therefore trained on SH coefficient decompositions of DTI from 100 WU-Minn HCP subjects (acquisition details in Section 2.2). For each subject, we calculate single-shell SH coefficient representations separately for each b-value shell (b=1000$\frac{s}{mm^2}$, b=2000$\frac{s}{mm^2}$, b=3000$\frac{s}{mm^2}$ independently), leading to 300 effective training samples. For each sample, we use a multi-channel volume with a concatenated mean low-b as the first channel, the zeroth-order SH representation as the second channel, and second-order SH representations as the last five channels. We refer to the augmented/degraded sample as the LR sample. We refer to the un-augmented sample used to compare the degraded sample to as the HR sample. During training, the neural network loss (described in detail in section 2.6) is based directly on the similarity between the HR sample and the superresolved LR sample.

\vspace{0.5cm}

We include multiple “angularly-invariant” augmentation steps to generate the LR sample for each training iteration. First, we randomly crop a $64\times64\times64$ voxel patch in the HR sample. Then, to mimic non-angularly dependent intensity fluctuations, we apply a random gamma field with a standard deviation of 0.1, as well as a random, smooth, low-frequency multiplicative bias field by sampling from a Gaussian distribution with $\sigma=0.2$ on a coarse (up to 4mm) grid to each low-b and zeroth-order channel of the HR sample. Finally, to generalize DiffSR across spatial resolutions and noise levels to avoid re-training on different DTI acquisition schemes, we generate the LR sample by injecting Gaussian noise with a maximum $\sigma=0.06$ across all channels, and applying Gaussian blurring followed by grid-wise resampling down to a spatial resolution randomly chosen between 1.5mm and 4mm. 

\vspace{0.5cm}

In addition to the geometric augmentation steps above, which do not rely on the angular structure of the HR sample, we apply aggressive “angular-aware” augmentation directly to SH coefficients during train-time through a domain-randomization strategy \citep{Tobin2017,Billot2023SS}. The first angular augmentation step we perform is SH rotation, which involves the following. For a unit direction vector $r\{\theta,\vartheta\} \in S^2$ defined along the polar angle $\theta$ and azimuthal angle $\vartheta$, let the complex SH basis for order $\ell$ and phase $m$ be denoted as: 

\begin{equation}
Y_{\ell}^{m}(\hat{r})
= (-1)^{m}\,\sqrt{\frac{2\ell+1}{4\pi}\,\frac{(\ell-m)!}{(\ell+m)!}}\;
P_{\ell}^{m}\!\bigl(\cos\theta\bigr)\,e^{i m \vartheta},
\end{equation}

where $P_{\ell}^{m}$ is the associated Legendre polynomial of degree $\ell$ and order m=-$\ell$,...,$\ell$. The raw radially defined diffusion signal $S(x,\hat{r})$ can be decomposed along an arbitrary $\hat{r}$ in terms of SH coefficients as: 

\begin{equation}
S(x,\hat{r}) = \sum_{\ell,m} c_{\ell}^{m}(x)\,Y_{\ell}^{m}(\hat{r}).
\end{equation}

We randomly rotate SH coefficients in the HR sample during training with Wigner D-Matrix operations \citep{vandeWiele2001}. To rotate a voxel consisting of SH coefficients $\mathbf{c}=\{c_{\ell}^{m}(x)\in\mathbb{C}\mid \ell=0\ldots l_{\max};\; m=-\ell\ldots \ell\}$ up to order ${\ell}_{max}$, with rotation matrix $\mathbf{R}=\mathbf{R}_z(\gamma)\mathbf{R}_y(\beta)\mathbf{R}_z(\alpha);\  \mathbf{R}\in \text{SO}(3)$ along a set of Euler angles $(\gamma,\beta,\alpha)$, one can solve: 

\begin{equation}
\mathbf{c}' = \bm{\mathcal{D}}(\mathbf{R})\,\mathbf{c},
\end{equation}

\noindent where

\begin{equation}
\bm{\mathcal{D}}(\mathbf{R}) = \mathrm{diag}\!\left\{\bm{\mathcal{D}}^{(0)}(\mathbf{R}),\ldots,\bm{\mathcal{D}}^{(l_{\max})}(\mathbf{R})\right\}.
\end{equation}

$\bm{\mathcal{D}}^{(\ell)}(\mathbf{R})\in\mathbb{C}^{N\times N}$ is a Wigner D-block for a single order $\ell$, which can be expressed in terms of elements of the reduced Wigner D-matrix based on the SO(3) Hermitian generator about the y-axis $J_y$, $\mathbf{d}^{(\ell)}_{m,m'}(\beta)=\langle \ell,m \mid \exp(-i\beta J_y)\mid \ell,m' \rangle,$ as:

\begin{equation}
\bm{\mathcal{D}}^{(\ell)}_{m,m'}(\bm{\mathcal{R}})
= \exp(-i m \alpha)\; \mathbf{d}^{(\ell)}_{m,m'}(\beta)\; \exp(-i m' \gamma).
\end{equation}

In practice, we use a real, second-order $\ell_{max}=2$ SH basis for $Y_{\ell}^m$ during neural network training and inference, and discard all odd orders of $\ell$ due to antipodal symmetry of the diffusion signal, and store only one complex/real sub-components of $\mathbf{c}$, as per convention in prior literature \citep{Tournier2019}. For the random SH rotation described above, we utilize the SHTools software package \citep{Wieczorek2018} for which we convert our real SH basis to a complex basis via zero-filling for compatibility. 

\vspace{0.5cm}

During training, we also deform randomly selected patches within the SH channels of the HR sample with a smooth displacement field from a coarse control grid of a random magnitude, which we up-sample with cubic interpolation. Let $\varphi(x) = x + \mathbf{u}(x)$ be defined as the spatial transform for a displacement \text{u}(x) at voxel x within a randomly chosen spatial patch $\Omega_u \subset \Omega$. Let the deformation gradient $\mathbf{F}(x)$ in terms of the Jacobian $\nabla \mathbf{u}(x)$ be defined as:

\begin{equation}
\mathbf{F}(x) = \nabla \varphi(x) = \mathbf{I} + \nabla \mathbf{u}(x), \qquad \forall x \in \Omega_u .
\end{equation}

Right polar decomposition yields the local rotation $\mathcal{R}(x) = \mathbf{F}(x)\mathbf{U}(x)^{-1}$ where $\mathbf{U}(x) = [\mathbf{F}(x)^{\mathsf T} \mathbf{F}(x)]^{\frac{1}{2}}$ is symmetric positive-definite. We reject displacement fields with a non-positive Jacobian determinant such that $\text{det}\{\mathbf{F}(x)\} \gtrapprox 0$ (with some minimal local folding permitted) as to maintain smooth, invertible, and orientation-preserving displacements, thus assuring that $\mathbf{U}$ is invertible and that $\mathcal{R} \in \text{SO}(3)$. Deformed second-order SH coefficients $\mathbf{c}_{def}$ are then calculated as: 

\begin{equation}
\mathbf{c}_{\mathrm{def}}(\varphi(x))
= \mathrm{diag}\!\left\{\mathbf{D}^{(0)}(\mathcal{R}(x)),\,\mathbf{D}^{(2)}(\mathcal{R}(x))\right\}\,\mathbf{c}(x),
\qquad \forall x \in \Omega_u .
\end{equation}

\noindent where, again, we restrict computation to real, odd-ordered $\mathbf{c}$. For patch-wise smoothness of the displacement field within the training volume and to maintain anatomic consistency, we enforce a Dirichlet boundary condition for $\Omega_u$ via cosine tapering of $\mathbf{u}(x)$ at the $\Omega_u$ boundary. Let $\eta(x) \equiv \text{dist}_{\infty}(x, \partial \Omega_u);\ x \in \Omega_u$ be the (internal) Chebyshev distance to the boundary of $\Omega_u$ and b be the tapering length. We define the modified displacement field $u_\eta (x)$ upon which we calculate the deformations in Eq. 15 and Eq. 16 as: 

\begin{equation}
\mathbf{u}_{\eta}(x)=
\begin{cases}
\mathbf{u}(x)\,\dfrac{1}{2}\!\left(1-\cos\!\left(\pi\,\dfrac{\eta(x)}{b}\right)\right),
& 0 \le \eta(x) < b,\\[0.6em]
\mathbf{u}(x),
& \eta(x) > d,
\end{cases}
\qquad \forall x \in \Omega_u .
\end{equation}

In the same deformed patch, we mimic local inconsistencies in SH coefficient estimates by applying a small SH “drift” to antipodal pairs of SH coefficients. We do this by first rotating the SH coefficients by a randomly chosen set of Euler angles, then multiplicatively applying a small intensity gain between uniformly chosen between 0.95 and 1.05 to either the $m \pm 1$ or the $m \pm 2$ antipodal SH pairs, then rotating back the SH coefficients with the same chosen Euler angles. This allows for modulation of single-channel SH coefficient intensities without affecting the overall signal rotation. 

\vspace{0.5cm}

In addition to explicit direction-dependent bias field correction that we describe in section 2.4, we also simulate random multiplicative bias fields directly in the HR sample SH domain at train-time to \textit{implicitly} teach the CNN any residual angular bias fields. We model these biases additively as: 

\begin{equation}
S_{\zeta}(x,\hat{r}) = S(x,\hat{r})\bigl(1+\zeta(x,\hat{r})\bigr),
\end{equation}

\noindent where the bias field $\zeta(x,\hat{r})$ (not to be confused with the modeled bias for Beta-DSW correction in section 2.4) can be expanded in the SH domain as:

\begin{equation}
\zeta(x,\hat{r}) = \sum_{\ell,m} \tau_{\ell}^{m}(x)\,Y_{\ell}^{m}(\hat{r}).
\end{equation}

\noindent leading to the expanded representation of the bias:

\begin{equation}
S_{\zeta}(x,\hat{r})
= \sum_{\ell,m} c_{\ell}^{m}(x)\,Y_{\ell}^{m}(\hat{r})
\;+\;
\sum_{\ell,m}\sum_{\ell',m'}
c_{\ell}^{m}(x)\,\tau_{\ell'}^{m'}(x)\,
Y_{\ell}^{m}(\hat{r})\,Y_{\ell'}^{m'}(\hat{r}).
\end{equation}

Using the SH product-to-sum identity via Gaunt integrals \citep{Politis2024,Homeier1996}, the harmonic basis can be simplified to:

\begin{equation}
Y_{\ell}^{m}(\hat{r})\,Y_{\ell'}^{m'}(\hat{r})
= \sum_{\ell'',m''} g_{\ell,\ell',\ell''}^{m,m',m''}(x)\,Y_{\ell''}^{m''}(\hat{r}).
\end{equation}

\noindent where $g_{\ell,\ell',\ell''}^{m,m',m''}(x)$ are Gaunt coefficients. By swapping the order of summation in Eq. 21, this leads to the updated form of the biased DTI signal:

\begin{equation}
S_{\zeta}(x,\hat{r}) = \sum_{\ell'',m''} c_{\ell''}^{m''(\zeta)}(x)\,Y_{\ell''}^{m''}(\hat{r}).
\end{equation}

\noindent where the synthetically biased signal coefficients $c_{\ell''}^{m''(\zeta)}(x)$ are expressed in terms of the original and biased coefficients:

\begin{equation}
c_{\ell''}^{m''(\zeta)}(x)
= c_{\ell''}^{m''}(x)
+\sum_{\ell,m}\sum_{\ell',m'}
g_{\ell,\ell',\ell''}^{m,m',m''}(x)\,c_{\ell}^{m}(x)\,\tau_{\ell'}^{m'}(x).
\end{equation}

In vector form, this can be interpreted as a simple linear mixing of $\mathbf{c}(x)$:

\begin{equation}
\mathbf{c}_{\zeta}(x) = \bigl(\mathbf{I} + \mathbf{M}\{\boldsymbol{\tau}(x)\}\bigr)\,\mathbf{c}(x).
\end{equation}

\noindent where $\mathbf{c}_{\zeta}(x)$ is the vector of synthetically biased signal SH coefficients, $\mathbf{M} \in \mathbb{R}^{6 \times 6}$ is a mixing matrix for real second-order ($\ell_{max}=2$) SH coefficients whose entries linearly depend on the bias coefficients $\tau_{\ell}^{m}(x)$. Rather than explicitly constructing $\mathbf{M}(x)$ and $\tau (x)$, which requires evaluating Gaunt groupings, we simulate direction-dependent bias at train-time with random low-rank (rank=2) mixing of $\mathbf{c}(x)$:

\begin{equation}
\mathbf{c}_{\zeta}(x) = \bigl(\mathbf{I} + \mathbf{V}(x)\mathbf{Q}(x)\bigr)\,\mathbf{c}(x),
\end{equation}

\noindent where $\mathbf{V}(x) \in \mathbb{R}^{6 \times 2}$ and $\mathbf{Q}(x) \in \mathbb{R}^{2 \times 6}$ are matrices where each entry is chosen uniformly between -0.025 and 0.025 as to minimize large fluctuations in SH coefficient values. 

\vspace{0.5cm}

Finally, we mimic a degraded angular resolution at train-time with mismatched augmentation between the HR and LR samples where the LR sample is subsampled to low angular resolution directly from the corresponding SH coefficients. We first generate a vertex set $\bm{\mathcal{F}}_{1} = \{\,f_{1,j}\mid j=1,\ldots,N_{1}\,\}$ from a regular icosahedron with a level-1 subdivision (containing $N_1$=42 total vertices) and project each vertex onto a unit 2-sphere, resulting in the unit-icosphere vertex set $\hat{\mathbf{u}} = \left\{\,\hat{\mathbf{u}}_{j}\;\middle|\; j=1,\ldots,N_{1},\ \hat{\mathbf{u}}_{j} = \frac{f_{1,j}}{\lVert f_{1,j} \rVert}\right\}$. Let the projection matrix $\mathbf{P} \in \mathbb{R}^{6 \times N_1}$ for all real second-order SH basis functions onto amplitudes discretized by $\hat{\mathbf{u}}$ be: 

\begin{equation}
P_{ij} = Y_{2}^{m_i}\!\left(\hat{\mathbf{u}}_{j}\right).
\end{equation}

\noindent and the projection of $\mathbf{c}(x)$ to the icosphere amplitude vector $\mathbf{h}(x) \in \mathbb{R}^{N_1}$ is therefore:

\begin{equation}
\mathbf{h}(x) = \mathbf{P}^{\mathsf T}\,\mathbf{c}(x).
\end{equation}

To mimic low angular resolutions, we choose a random subset of amplitude vertices on the icosphere and subsequently add additive noise to the non-zero amplitudes such that, where the subsampled icosphere projections $\tilde{\mathbf{h}}(x)$ can be expressed as:

\begin{equation}
\tilde{\mathbf{h}}(x) = \mathbf{S}\,\mathbf{h}(x) + \varepsilon_{p},
\end{equation}

\noindent where $\mathbf{S}$ is the row sub-selection matrix with between four and nine non-zero rows are randomly chosen at every training iteration and $\varepsilon_{p} \sim \mathcal{N}(0,\mathbf{I}\sigma^2)$ with $\sigma=0.02$. Because deprojection of $\tilde{\mathbf{h}}(x)$ back to SH space via a pseudoinverse of $\mathbf{P}$ is underdetermined, we recover the angularly subsampled SH coefficients $\tilde{\mathbf{c}}(x)$ from icosphere amplitude space via ridge regression:

\begin{equation}
\tilde{\mathbf{c}}(x)
= \arg\min_{\mathbf{c}(x)}
\left\{
\left\lVert \mathbf{P}^{\mathsf T}\mathbf{S}\mathbf{c}(x) - \tilde{\mathbf{h}}(x) \right\rVert_2^{2}
+ \lambda \left\lVert \mathbf{c}(x) \right\rVert_2^{2}
\right\},
\end{equation}

\noindent which leads to the analytic solution:

\begin{equation}
\tilde{\mathbf{c}}(x)
=
\left[\bigl(\mathbf{P}^{\mathsf T}\mathbf{S}\bigr)^{\mathsf T}\bigl(\mathbf{P}^{\mathsf T}\mathbf{S}\bigr) + \lambda \mathbf{I}\right]^{-1}
\bigl(\mathbf{P}^{\mathsf T}\mathbf{S}\bigr)^{\mathsf T}\,\tilde{\mathbf{h}}(x).
\end{equation}

When called, $\tilde{\mathbf{c}}(x)$ replaces $\mathbf{c}(x)$ in the LR sample in a training batch while maintaining the same HR sample.

\subsection{Neural network architecture and training details}

Prior to spatial superresolution, we impart angular correspondence with dependence on angular proximity on the SH coefficients by first projecting the SH inputs c onto an icosahedral surface, as in section 2.5, followed by de-projection back to SH space. We assume that, unlike in SH space, mixing of angular information is most informative within geodesic proximity on the icosphere. To capture this dependence in a compact manner, we apply two graph-convolution layers to $\mathbf{h}$ where we define the graph connectivity as restricted to the set of K-nearest neighbors (K=6) to each vertex in $\hat{\mathbf{u}}$. Specifically, we define the adjacency matrix  $\mathbf{A} \in \mathbb{R}^{N_1 \times N_1}$ for vertex $\hat{\mathbf{u}}_i$ as:

\begin{equation}
A_{ij} =
\begin{cases}
1, & \hat{\mathbf{u}}_j \in \mathrm{KNN}(\hat{\mathbf{u}}_i),\\
0, & \text{otherwise}.
\end{cases}
\end{equation}

\noindent where $\mathrm{KNN}(\hat{\mathbf{u}}_i)$ is the K-nearest neighbors set for $\hat{\mathbf{u}}_i$. For voxel $v$, we define the amplitude feature tensor $\mathbf{H}_v \in \mathbb{R}^{N_1 \times N_f}$ where $N_f$ is the input layer feature dimensionality. The output tensor $\mathbf{H}_v^{out} \in \mathbb{R}^{N_1 \times N_{f'}}$, where $N_{f'}$ is the output layer feature dimensionality, is calculated for a single graph-convolution layer as:

\begin{equation}
\mathbf{H}^{\mathrm{out}}_{v}
=
\mathrm{GELU}\!\left(
\mathrm{LN}\!\left\{
\left(w_{v}\mathbf{I} + \frac{w_{n}}{\lvert \mathrm{KNN}\rvert}\,\mathbf{A}\right)\mathbf{W}\,\mathbf{H}_{v}
+ \mathbf{1}\mathbf{b}^{\mathsf T}
\right\}
\right).
\end{equation}

\noindent where $w_v,\ w_n \in \mathbb{R}$ are learned vertex/neighborhood mixing scalars, $\mathbf{W} \in \mathbb{R}^{N_f \times N_{f'}}$ and $\mathbf{b} \in \mathbb{R}^{ N_{f'} \times 1}$ are learned layer weights and (broadcast) biases, $\text{LN}\{\cdot\}$ is layer normalization, and ${\lvert \cdot\rvert}$ is the set cardinality. The filtered amplitude output from the graph-convolution layers is then re-projected back into SH space with a Moore-Penrose pseudo-inverse of ${\mathbf{P}}$.

\vspace{0.5cm}

\begin{figure}[!htbp]
    \centering
    \includegraphics[width=0.99\linewidth]{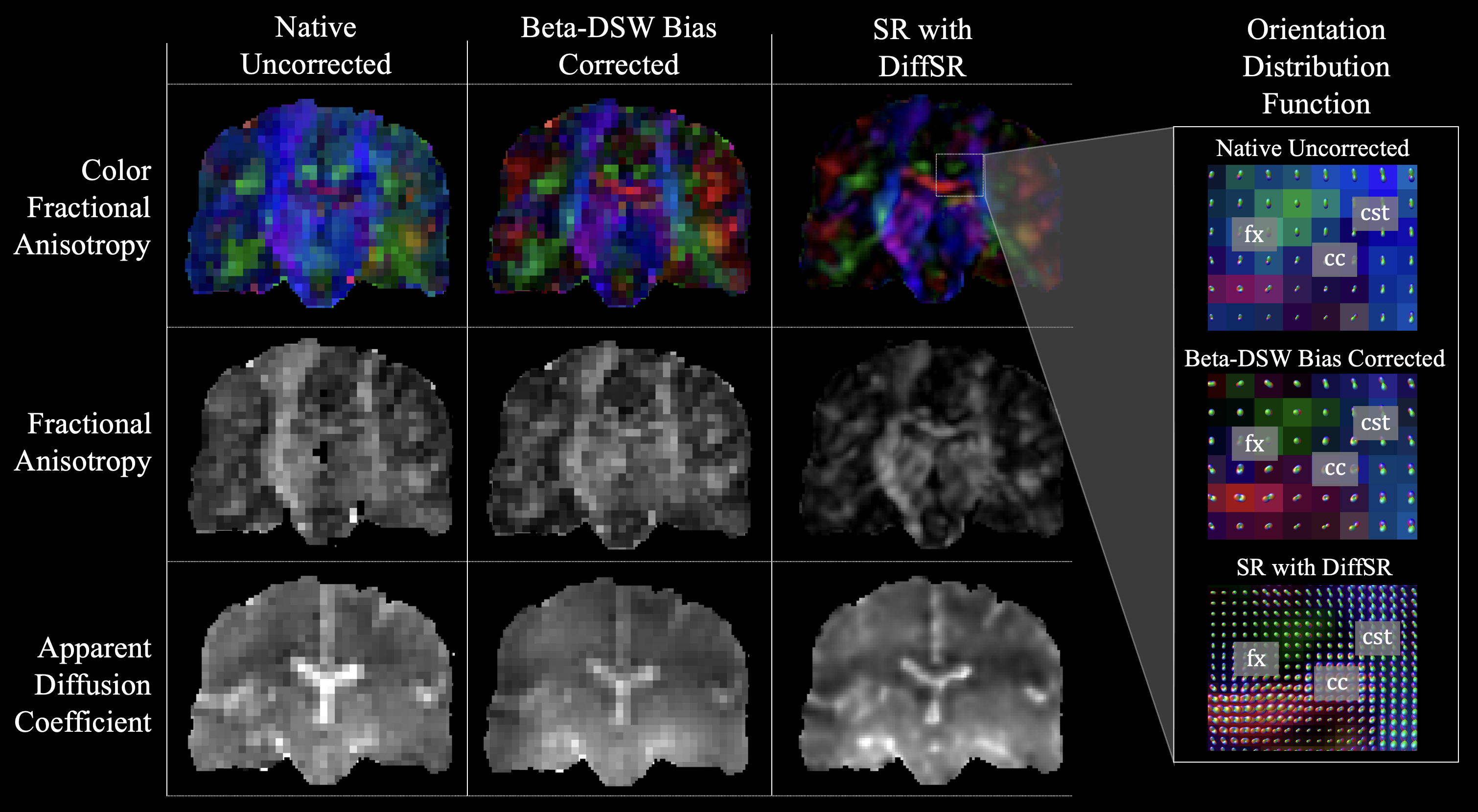}
    \caption{\textbf{Graphical overview of direction-specific bias correction followed by superresolution with DiffSR applied to ULF DTI variants}. Shown in the left column are the native, uncorrected color FA, FA and ADC coronal views of a representative ULF DTI scan from our custom dataset. In the middle column are coronal views of the same subject with Beta-DSW bias correction applied. In the right column are the superresolved outputs (with DiffSR applied to the bias corrected sequence). We also show zoomed-in views of the second-order orientation distribution functions calculated for each processing step at the corpus callosum-corticospinal tract junction. cst: corticospinal tract, fx: fornix, cc: corpus callosum, SR: superresolution.}
    \label{fig:Figure_4}
\end{figure}

We feed the low-b channel concatenated with the SH output of the “icosphere projection-deprojection” block (for a total of 7 channels: low-b + SH coefficients) through a 3-dimensional U-Net CNN model \citep{Ronneberger2015} that maps to a 7-channel output with the same channel configuration. The U-Net contains four encoder and decoder layers with a base layer number of 128 features. For each layer, we utilize $3 \times 3 \times 3$ convolution kernels with GELU activation functions, with mean-pooling for downsampling between encoder layers and $2 \times 2 \times 2$ transposed convolutions with a stride length of two voxels for upsampling between decoder layers. To allow the network to condition decoder upsampling on global features, we implement an attention block with eight learnable global tokens at the U-Net bottleneck. Finally, to re-capture angular dependencies between processed SH coefficients, the SH output channels of the U-Net are passed through a second icosphere projection-deprojection block. A graphical overview of the entire DiffSR pipeline is shown in Figure $\ref{fig:Figure_3}$. An example the application of Beta-DSW bias correction followed by superresolution with the DiffSR pipeline is shown in Figure $\ref{fig:Figure_4}$.

\vspace{0.5cm}

We train DiffSR for 1500 epochs with 20 iterations of backpropagation per epoch and a batch size of four samples. During training, we use ADAM optimization with betas of 0.9 and 0.95 and an initial learning rate of $10^{-4}$ with linear warm-up from $10^{-5}$ for 100 epochs. We utilize a channel-wise loss function, which is split into L2 loss for the mean low-b and $\ell=0$ SH channels (with a weighting of 5), and L1 loss for the $\ell=2$ channels (with a weighting of 10). To directly penalize discrepancies between the major signal-space angles generated by the SH coefficients, we also incorporate an angular loss (with a weighting of 1), which is the mean squared angular error between differentiable principal directions in the LR and HR sample SH coefficients, which are calculated using a soft argmax across a predefined set of angular directions on the surface of a sphere (obtained through Fibonacci sampling). Finally, we regularize the training penalty with a forward model-based loss meant to maintain the anatomic consistency and fidelity of SR (with $\lambda$=2.5). Specifically, we apply spatial downsampling with the same Gaussian blurring and resampling parameters (see section 2.5) as used in the concurrent training iteration to the output SR prediction of the LR sample, and calculate the mean squared error of this resampled output with respect to the original LR sample.

\vspace{0.2cm}

\vspace{-0.2cm}

\section{Results}

\subsection{Reconstruction accuracy with synthetically-downsampled HCP DTI data}

The Connectom HCP dataset (acquisition details in Section 2.2) dataset (1) possesses a field strength and spatial resolution that is within the ideal target range of DiffSR and (2) is sufficiently different in terms of scanner and sequence type to the WU-Minn HCP dataset used to train DiffSR, as to directly test its generalizability. The raw low-b sequence and diffusion-encoding gradient directions were synthetically downsampled with the MRtrix mrgrid command and trilinear interpolation with to generate low-resolution inputs. We chose a target downsampled resolution range of 2-4 mm in 0.25mm increments. To simulate low angular resolutions, DiffSR reconstructions were performed on random n-subsets of gradient directions with downsampling factor (the ratio of the number of target directions to the original number of directions per-shell) between 0.15 and 0.55. Direction subsets were chosen with greedy furthest-point electrostatic repulsion as to cover the unit sphere as uniformly as possible and to preserve antipodal pairs as to retain an unbiased diffusion tensor reconstruction. We chose voxel-wise mean absolute error (MAE) and local (window size of 10 voxels) normalized cross correlation (LNCC) for the SH coefficient channels, FA, and ADC reconstructions to capture scalar diffusion metric error and overall contrast profile fidelity, and voxel-wise mean absolute angular error for the V1 reconstructions to assess the net angular accuracy of DiffSR SH reconstructions. We evaluated DiffSR reconstructions on the b=1000/3000/5000$\frac{s}{mm^2}$ shells separately to test any b-value-dependence in terms of reconstruction accuracy. Reconstruction accuracy for the b=1000 s/mm2 shell (which is the b-value closest to our ULF DTI acquisition) is shown in Figure $\ref{fig:Figure_5}$. Reconstruction accuracy for the b=3000/5000$\frac{s}{mm^2}$ shells are provided in Supplementary Figure $\ref{fig:suppF1}$. Overall, DiffSR reconstructions provided greater degrees of accuracy and contrast recovery in terms of both MAE and LNCC for both SH coefficients and FA up to spatial resolutions of around 3mm and across all angular resampling factors with the exception of the lowest angular downsampling factor (0.15: $\sim$9 direction subset), where DiffSR outperformed trilinear upsampling up to spatial resolutions of $\sim$2.25mm. DiffSR showed superior V1 reconstructions in terms of angular error up to resolutions of $\sim$3.5mm and across angular downsampling factors. However, DiffSR generated relatively poor ADC reconstructions, with it outperforming trilinear upsampling only until spatial resolutions of $\sim$3.75mm in terms of MAE and displaying worse performance than trilinear upsampling for all spatial and angular resolutions in terms of LNCC. A similar accuracy pattern was observed for the b=3000/5000 s/mm2 shells, albeit with poorer accuracy increases for DiffSR in the b=5000$\frac{s}{mm^2}$ shell, especially for ADC similarity, as shown in Supplementary Figure $\ref{fig:suppF1}$.

\begin{figure}[t]
    \centering
    \includegraphics[width=0.99\linewidth]{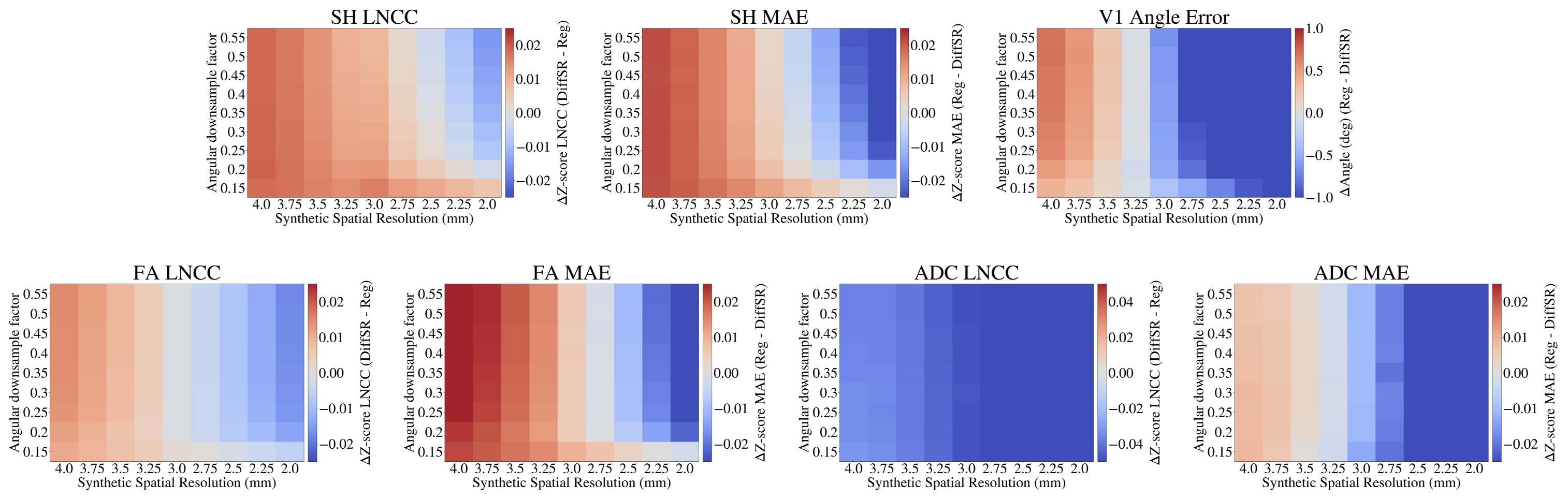}
    \caption{\textbf{DiffSR reconstruction accuracy under synthetic spatial and angular downsampling}. DiffSR reconstruction accuracy is shown with respect to standard trilinear upsampling for the b=1000$\frac{s}{mm^2}$ shell of synthetically downsampled HF DTI data from the Connectom HCP dataset. The raw (i.e., gradient directions and low-b) HF DTI data were spatially downsampled with trilinear interpolation between 2mm and 4mm isotropic spatial resolutions at 0.25mm intervals. The data was also angularly downsampled by choosing random gradient direction subsets at ratios of 0.15 to 0.55 with respect to the original gradient number in the respective shell. Shown are the MAE and LNCC for the SH coefficient channels (channel-wise average), FA and ADC reconstructions. Also shown is the voxel-wise mean angular error for the V1 reconstructions.}
    \label{fig:Figure_5}
\end{figure}

\subsection{Superresolution of degraded DTI data for white matter analysis in Alzheimer's disease}

White matter analysis in AD/MCI using ULF MRI is becoming increasingly popular due to its portability and ease of use in outpatient settings \citep{Sorby-Adams2024,Farnan2025}. However, as previously described, ULF DTI-related signal degradation hinders the analysis of individual white matter tracts. We test the capability of DiffSR in superresolving degraded tract-specific microstructural information relevant to AD white matter pathophysiology in ADNI control and AD/LMCI groups (see Section 2.2). To mimic the parameters of our ULF DTI sequence, we downsampled the original ADNI DTI sequence to 3.5mm isotropic spatial resolution, chose random subsets of 9 diffusion-encoding gradient directions with greedy furthest-point electrostatic repulsion, and injected Rician noise in the raw diffusion signal space with a noise standard deviation of 100. We henceforth call this the “degraded” DTI sequence. We subsequently ran DiffSR on the degraded DTI sequence and compared group changes in individual white matter tract FA and ADC in both the degraded and superresolved DTI with respect to the original DTI sequence. We used the Benjamini-Hochberg false discovery rate (FDR)-corrected two-tailed Wilcoxon rank-sum test \citep{Benjamini1995} to determine statistical significance for group-wise changes in tract FA/ADC.

\vspace{0.5cm}

\begin{figure}[!htbp]
    \centering
    \includegraphics[width=0.98\linewidth]{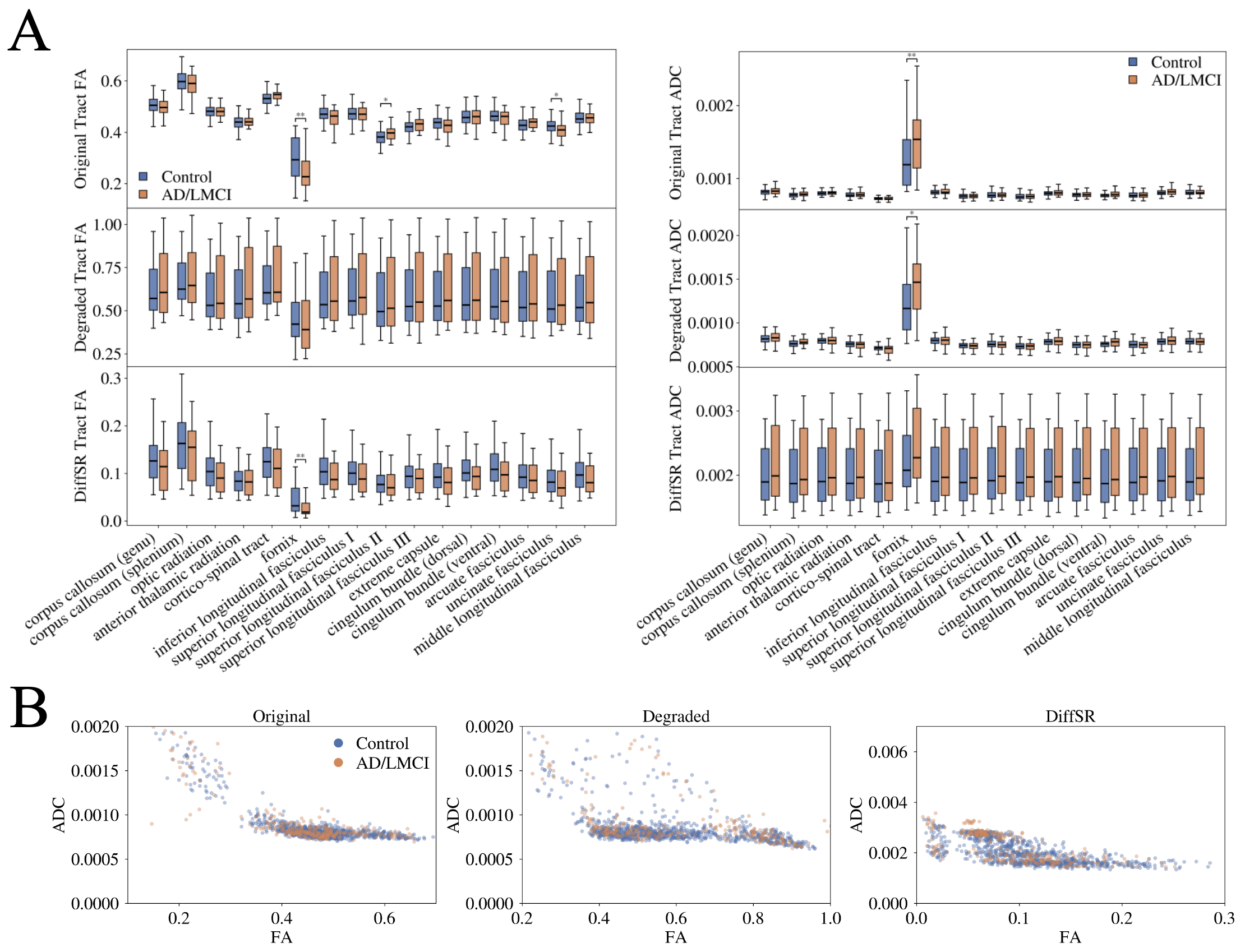}
    \caption{\textbf{Per-tract fractional anisotropy and apparent diffusion coefficient measurements across ADNI control and AD/LMCI subjects}. Shown in \textbf{(A)} are the tract-wise distributions of tract-averaged FA (left panel) and ADC (right panel) for original, degraded, and superresolved DTI reconstructions of ADNI control (blue) and AD/LMCI (orange) subject groups. Significance bars indicate Benjamini Hochberg FDR-corrected two-tailed Wilcoxon rank-sum test p-values of <0.05 (*) and <0.01 (**) respectively. Shown in \textbf{(B)} are joint FA and ADC values for both control and AD/LMCI subjects for original, degraded, and superresolved DTI reconstructions plotted along orthogonal FA and ADC axes as to visualize clustering and spread for both groups.}
    \label{fig:Figure_6}
\end{figure}

Super-resolved degraded DTI with DiffSR most notably showed tract-wise group decreases in FA between control and AD/LMCI groups corresponding to similar reductions in the original DTI scans, as seen in Figure $\ref{fig:Figure_6}$A. These FA reductions were not observed in the degraded DTI scan, which showed no statistically significant tract-wise FA group changes. DiffSR recovered tract-wise FA reduction in the fornix (p<0.01), which has shown significant DTI-based changes in prior AD literature \citep{Nir2013,Oishi2014} and was the tract with the most significant group-wise FA reduction in the original DTI sequence (p<0.01). Furthermore, while not deemed statistically significant, DiffSR showed notable FA reduction in other tracts with significant AD association, most notably the uncinate fasciculus and inferior longitudinal fasciculus \citep{Sexton2011,Zhang2009} (uncorrected p-value: 0.05, FDR corrected p-value: 0.18 for both tracts), which were significant or near-significant in terms of FA reduction for the original DTI sequence (respective uncorrected p-values: <0.01/0.01, FDR corrected p-values: 0.02/0.07). FA reduction was not observed for either tract in the degraded DTI scan (respective uncorrected p-values: 0.59, FDR corrected p-values: 0.66 for both tracts). For ADC, while overall group-wise increases were observed with DiffSR, as consistent with prior literature and observed in both the original and degraded DTI scans, no ADC increases were statistically significant. This included ADC increases in the fornix, which were both deemed statistically significant in the original DTI sequence (p<0.01) and degraded DTI sequence (p<0.01). All uncorrected and FDR corrected p-values for each tract can be found in Supplementary Tables $\ref{tab:suppT2}$ and $\ref{tab:suppT3}$ respectively. Finally, we jointly assessed the individual per-tract and per-subject FA and ADC values along orthogonal FA and ADC spatial axes for all control and AD/LMCI subjects, as seen in Figure $\ref{fig:Figure_6}$B. Both the original and DiffSR FA/ADC scatters showed more visually distinct control and AD/LMCI clusters (mainly along the FA axis for the original DTI scatter, and along both FA and ADC axes for the DiffSR scatter), while the degraded DTI scatter displayed slightly more spatial mixing between the two respective clusters. To quantify this group separation in white matter tracts with known DTI-related changes in AD, we constructed minimal linear discriminant scores by projecting individual FA/ADC points from the genu/splenium of the corpus callosum, dorsal/ventral cingulum bundle, fornix, superior longitudinal fasciculus (\textit{Tracula} subunits 1-3 combined), inferior longitudinal fasciculus, and uncinate fasciculus \citep{Nir2013,Sexton2011,Zhang2009,Oishi2014} onto a Fisher linear discriminant direction (underlying a linear discriminant analysis projection), which we estimated via leave-one-out cross-validation separately for the original, degraded, and DiffSR scatters. We intentionally used a minimal linear projection to avoid imposing a complex model and learned hyperparameters as to assess the scatter geometry in FA/ADC space in a simple fashion. Using the area under the receiver operating characteristic curve (AUC) to quantify group separation, DiffSR yielded the highest discrimination (AUC=0.59), as compared to the original (AUC=0.54) and degraded (AUC=0.55) DTI scatters. The difference between the DiffSR and degraded AUCs was statistically significant (two-sided p-value=0.03) based on a paired bootstrap of AUC differences over subjects (resampling subjects with replacement while keeping all tract-wise FA/ADC estimates within each subject), as within-subject tract-wise FA/ADC are assumed to be correlated. These results potentially suggest that DiffSR enhances FA/ADC-based control/AD classification at clinical-grade and/or degraded DTI resolutions. 

\subsection{Comparisons with matched high-field DTI}

To test the fidelity of both Beta-DSW bias field correction and superresolution with DiffSR, we compared the per-subject FA, ADC and V1 coherence in \textit{Tracula}-segmented white matter tracts between ULF DTI scans and matched HF DTI scans in our ULF dataset. Because we do not have perfect alignment between HF and ULF scans for each participant, rather than performing voxel-wise similarity assessment of directionality, we took a per-tract measurement of V1 coherence, $\text{VC} \in \mathbb{R}$, which we define as:

\vspace{-0.5cm}

\begin{equation}
\mathrm{VC}
=
\left\lVert
\frac{1}{N_t}\sum_{x=1}^{N_t}
\operatorname{sign}\!\bigl(\mathbf{v1}(x)^{\mathsf T}\hat{\mathbf{r}}\bigr)\,\mathbf{v1}(x)
\right\rVert_2^{2},
\end{equation}

\vspace{0.3cm}

\noindent where $\text{sign}(\cdot)$ is the sign function, $\hat{r}$ is a reference unit direction, which we assign as the mean V1 within the respective tract ROI, $N_t$ is the total number of voxels for the tract ROI, and $\left\lVert \cdot \right\rVert_2^{2}$ is the 2-norm. As such, V1 coherence measures the overall homogeneity in directionality within a tract ROI (i.e., in “straight” tracts with dominant and unchanging directionality, $\text{VC} \approx 1$, and in highly tortuous tracts with changing directionality, $\text{VC} \approx 0$). For quantification of HF-ULF agreement, we assessed the inter-tract intraclass correlation coefficients (ICC) (two-way random effects, absolute agreement) between HF DTI and native, uncorrected ULF DTI with only standard preprocessing (see section 2.2), Beta-DSW-corrected ULF DTI, and finally DiffSR-superresolved on both native and Beta-DSW-corrected ULF DTI data (henceforth referred to as “superresolved”). To correct for any global intensity shifts, we z-scored all tract-wise FA and ADC measurements (i.e., across all segmented tracts) for ICC estimates and overall visualization.

\vspace{0.5cm}

\begin{figure}[!htbp]
    \centering
    \includegraphics[width=0.96\linewidth]{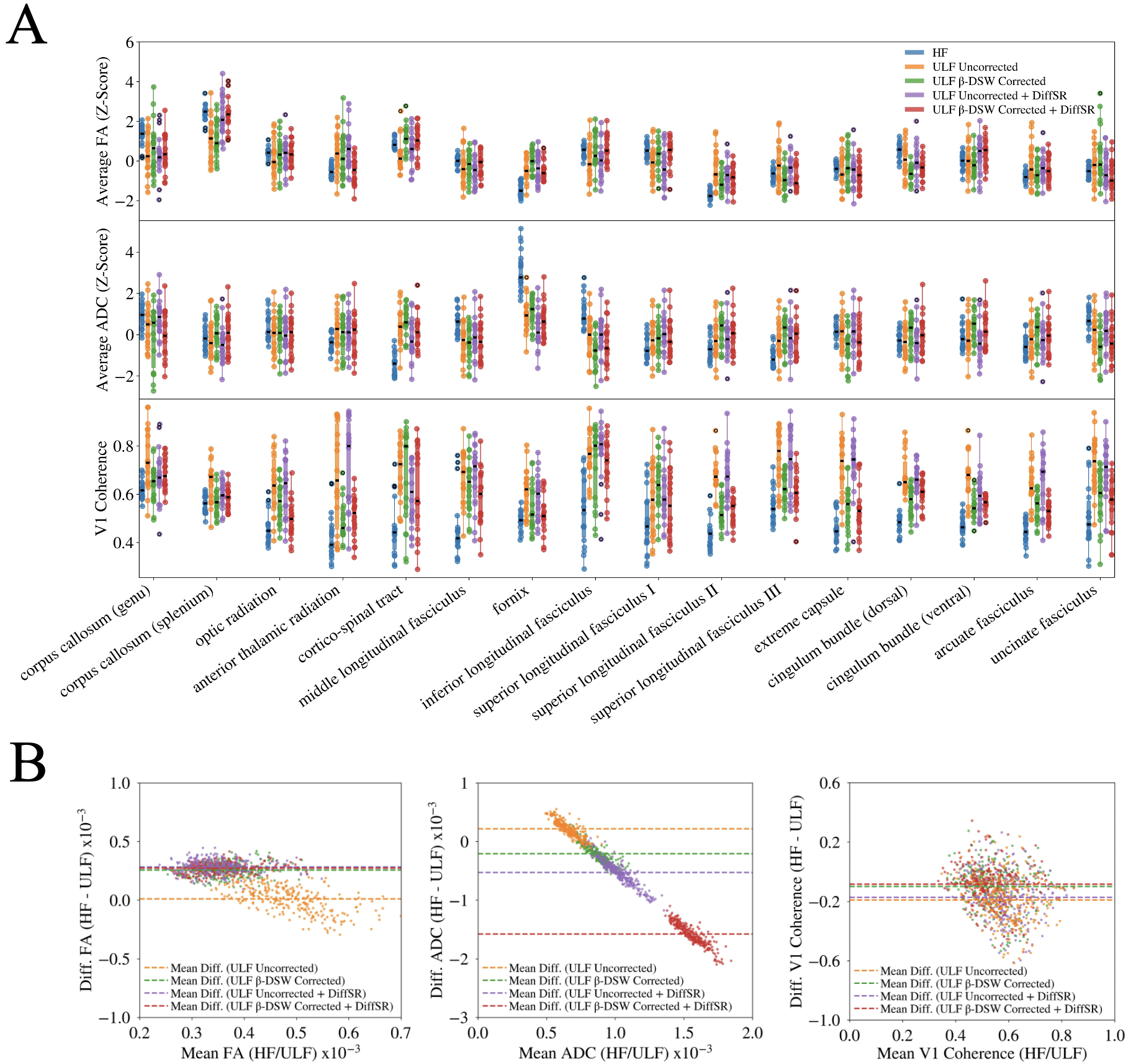}
    \caption{\textbf{Per-tract z-scored fractional anisotropy and apparent diffusion coefficient, as well as principal eigenvector coherence measurements across ULF DTI reconstruction variants}. Shown in \textbf{(A)} are z-scored distributions of tract-averaged FA (top), ADC (middle), and V1 coherence (bottom) for our 18-subject cohort with matched conventional HF DTI and ULF DTI sequences across a subset of white matter tracts segmented with \textit{Tracula}. The native ULF DTI sequence with standard preprocessing (orange) is compared to Beta-DSW bias-corrected ULF DTI (green) ULF DTI, and to DiffSR applied to uncorrected (purple) and Beta-DSW bias corrected ULF DTI (red) in terms of overall agreement with matched HF DTI measurements (blue). Shown in \textbf{(B)} are Bland-Altman plots for all non-z-scored ULF DTI per-tract FA (left), ADC (middle), and V1 coherence (right) measurements as compared to the matched HF DTI.}
    \label{fig:Figure_7}
\end{figure}

\begin{figure}[!htbp]
    \centering
    \includegraphics[width=0.999\linewidth]{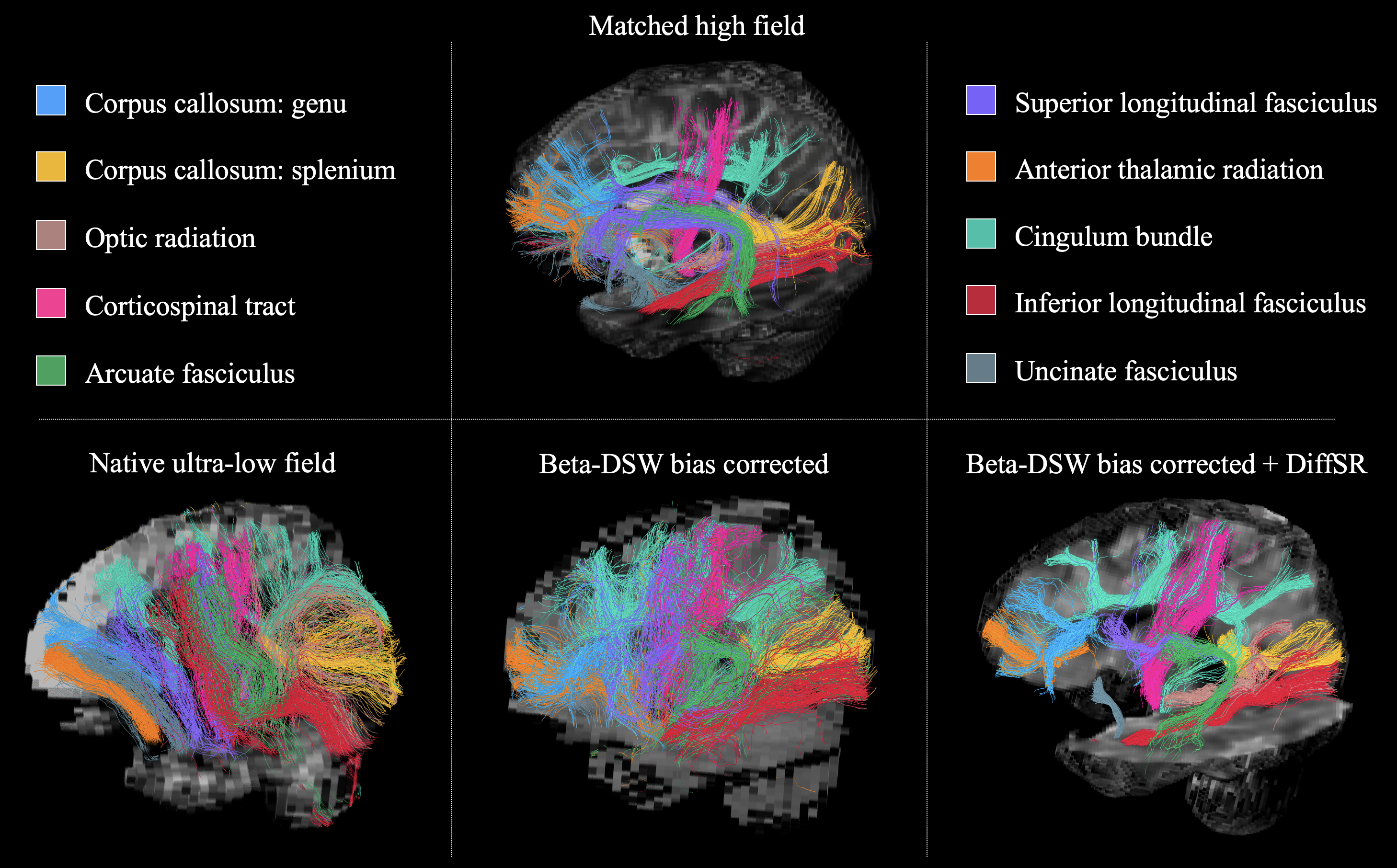}
    \caption{\textbf{Deterministic tractography applied to ULF DTI postprocessing variants}. In the top row are deterministic tractography reconstructions shown over a mean low-b volume for a set of \textit{Tracula}-derived white matter tracts in a HF DTI acquisition from a representative subject in our ULF dataset. In the bottom row are corresponding deterministic tractography reconstruction in the native (left), Beta-DSW bias corrected without (middle), and with application of DiffSR (right). Deterministic streamlines were generated with the MRtrix tckgen command using the iFOD2 algorithm and propagated through pre-defined \textit{Tracula} white matter segmentation cross-sections. The superior longitudinal fasciculus streamlines shown are the union of the \textit{Tracula} “slf2” and “slf3” ROIs, which is a commonly derived delineation of the tract \citep{Janelle2022}.}
    \label{fig:Figure_8}
\end{figure}

Upon inspection of the non-z-scored FA and ADC measurements (Supplementary Figure $\ref{fig:suppF2}$), native ULF DTI showed a global decrease in both FA and ADC as compared to matched HF DTI. V1 coherence was notably increased in all ULF DTI reconstructions as compared to HF DTI (Figure 7A). Native ULF DTI showed a global reduction in per-tract variance amongst subjects for FA, and a global increase in variance for ADC and V1 coherence. Beta-DSW bias-corrected ULF DTI resulted in further global FA reduction, but a global increase in ADC. Both DiffSR ULF DTI reconstructions displayed the greatest global reduction in both the mean and variance for FA, and the greatest global increase for ADC. DiffSR on the Beta-DSW bias corrected ULF DTI showed the greatest reduction in variance for V1 coherence. Inspection of the z-scored DTI data (Figure $\ref{fig:Figure_7}$A) showed that for most tracts with notably low (e.g., fornix and anterior thalamic radiation) or high (e.g., splenium of the corpus callosum), mean FA values for both superresolved ULF DTI reconstructions were the closest overall to the HF DTI mean FA. However, native ULF DTI showed the greatest overall correspondence in mean FA to HF DTI in terms of the mean tract-wise ADC values. These observations are largely corroborated by ICC measurements. For FA, both superresolved ULF DTI reconstructions (ICC=0.86 for Beta-DSW bias corrected DiffSR, ICC=0.73 for uncorrected DiffSR) and Beta-DSW bias corrected ULF DTI (ICC=0.68) showed notably greater agreement with HF DTI than native ULF DTI (ICC=0.56). For ADC, native ULF DTI, overall agreement with HF DTI was poorer, with native ULF DTI showing the greatest agreement (ICC=0.37) and Beta-DSW bias correction resulting in the least agreement (ICC=0.05). Superresolved uncorrected ADC showed slightly worse agreement as compared to native ULF DTI (ICC=0.27). Beta-DSW bias corrected and superresolved ADC displayed a slight jump in agreement (ICC=0.13) as compared to Beta-DSW bias correction alone. For V1 coherence, both superresolved Beta-DSW bias corrected ULF DTI (ICC=0.34) and Beta-DSW bias corrected ULF DTI (ICC=0.20) showed notably greater agreement with HF DTI than either native (ICC=0.06) or superresolved uncorrected (ICC=0.02) ULF DTI. A Bland-Altman plot of the non-z-scored metrics as compared to matched HF DTI measurements (Figure $\ref{fig:Figure_7}$B) revealed a bias that was linearly dependent on the magnitude of both the net anisotropy (FA) and overall diffusivity (ADC) across all ULF DTI reconstructions. For FA measurements, Beta-DSW bias corrected ULF DTI and both superresolved ULF DTI reconstruction variants both displayed reduced dispersions and reduction in linear dependence on mean FA, but the inclusion of a persistent positive mean bias as compared to native ULF DTI, consistent with systematically lower non-z-scored FA as compared to HF DTI. For ADC measurements, all ULF DTI reconstructions show comparable dispersion but exhibited a proportional bias that varied linearly with the mean ADC, suggesting a presence of a calibration or effective diffusion weighting (i.e., b-value) mismatch. All three ULF DTI reconstructions showed a negative-linear relationship with mean ADC, but the magnitude of this bias was most pronounced in each superresolved ULF DTI reconstruction variant (mean offset of $-0.53*10^{-3}$ for uncorrected DiffSR and $-1.6*10^{-3}$ for Beta-DSW bias corrected DiffSR). For V1 coherence, Bland-Altman plotting indicated a negative bias for native ULF DTI, which was slightly attenuated following Beta-DSW correction and both DiffSR variants. Finally, inspection of deterministic tractography in a representative subject from the ULF DTI dataset (Figure $\ref{fig:Figure_8}$) as compared to matched HF DTI showed that Beta-DSW bias correction resulted in more significantly consistent tract morphology and overall streamline geometry for large supra-tentorial tracts such as the corpus callosum, corticospinal tracts and optic radiation. Furthermore, the application of DiffSR to the Beta-DSW bias corrected ULF DTI resulted in further refinement of streamline geometry in supra-tentorial tracts, especially in tracts with thinner cross-sectional areas such as the arcuate fasciculus and portions of the cingulum bundle.

\vspace{0.9cm}

\vspace{-0.2cm}

\newpage

\section{Discussion}
In this work, we developed a 9-direction DTI sequence adapted to ULF scanning systems and companion enhancement algorithms that combine Q-space-aware Bayesian artifact correction built on Beta and DSW distribution priors coupled with spatio-angular superresolution, which we term DiffSR. Across our synthetic spatial and angular degradation experiment in HCP data, we observed consistent accuracy gains in both raw spherical harmonic coefficients and DTI-derived metrics with the application of DiffSR as compared to the un-degraded, ground-truth data. We proceeded to show that application of DiffSR improves DTI-derived contrast in the form of FA and ADC within key white tracts in AD/LMCI data under ULF-level synthetic signal degradation, supporting the potential utility of DiffSR beyond ULF and in DTI sequences with spatial/angular resolution and SNR constraints, as common in a clinical setting. Finally, we demonstrate through tract-level microstructural comparisons in a custom ULF DTI dataset with matched HF DTI that dominant ULF failure modes that are not fully addressed by conventional preprocessing directly benefit from a direction-aware bias correction model coupled with DiffSR.  Together, these results suggest that portable ULF DTI, despite severe hardware-level constraints resulting in imaging artifacts, poor SNR and resolution constraints, can be made more clinically viable with aggressive correction and superresolution algorithms. 

\vspace{0.5cm}

From an artifact-correction standpoint, we observe and specifically tackle the correction of direction-dependent bias fields that occur in ULF DTI. In standard DTI sequences, standard bias correction algorithms extrapolate bias fields from low-b images which ignores any Q-space dependence. In ULF DTI, Q-space dependent bias fields are a significant contributor to overall artifacting, as evidenced by prior literature \citep{Gholam2025,Ametepe2024}. We therefore provide an adaptive Bayesian correction method that jointly relies on the directionality of the underlying diffusion tensor through the estimated V1, as well as its anisotropy through FA. While standard DTI bias correction does not take directionality into account, we group acquisition-induced artifacting into direction-dependent multiplicative fields that are estimated within a generative model through conditioning of a DSW prior calculated from V1. By directly writing the likelihood in terms of V1, our model relies on a stable statistic of the diffusion tensor, which as a result makes for simple/stable optimization and invariance to the exact number of diffusion-encoding directions. To regularize the DSW likelihood, we also incorporate a Beta distribution prior on the underlying tissue FA. Such a prior provides a tissue-aware expectation for how much directional correction is plausible locally, which disambiguates true bias fields from microstructure affecting tissue contrast. That is, in high-FA white matter bundles, the bias correction model preserves orientation-dependent contrast that is consistent with high degrees of anisotropy and avoids overcorrection of signal intensities in those regions \citep{Iglesias2019}. Conversely, in regions with low expected anisotropy (such as cortical gray matter and CSF), the same Beta prior encourages the model to prioritize absorbing direction-specific intensity differences into the calculated bias fields. This behavior can be visualized in Figure $\ref{fig:Figure_2}$. 

\vspace{0.5cm}

The effect of the Beta-DSW bias correction on the removal of direction-dependent bias fields is evident qualitatively in our ULF DTI dataset, which resulted in visually significant reversion of directionality of diffusion to anatomically plausible orientations within key white matter pathways. This included (overall) superior-inferior directionality in the corticospinal tracts or left-right directionality within the corpus callosum. This is further evidenced by more visually plausible deterministic tractography reconstructions of such major white matter pathways, as seen in the color FA plots in Figure 4. Of note, to avoid over-correction of the directionality of diffusion and simply overfit to an “HCP-like” target, we model bias fields with low-order, spatially smooth basis functions and regularize the magnitude of their respective coefficients. Quantitatively, Beta-DSW bias correction yielded a significant increase in agreement (in terms of ICC) with matched HF DTI as compared to uncorrected ULF DTI. As such, these results indicate that this bias correction method yields accurate, robust and acquisition-agnostic bias field estimates that are adaptive to the underlying tissue class and reflect artifacting common to ULF DTI space. 

\vspace{0.5cm}

Even with correction of direction-dependent bias fields, ULF DTI data retain residual non-smooth artifacts and noise that are difficult to model explicitly and fall outside of the assumptions made by standard HF preprocessing, correction or superresolution methods. More specifically, state-of-the-art DTI superresolution algorithms which utilize methods ranging from compressed-sensing \citep{Michailovich2011} and structural similarity indices \citep{Coup2013} to deep learning-based models acting on ODF representations \citep{Zeng2022} generally assume HF SNR, denser q-space sampling schemes, and generally more stable acquisitions. This motivates our second contribution, DiffSR, which addresses this gap for ULF DTI data by implicitly correcting for residual artifacting and abnormally low SNR with sparse angular sampling schemes through aggressive augmentation in SH space coupled with a neural network architecture that explicitly imposes consolidating angular information. For train-time augmentations, we specifically focus on simulating ULF-specific degradation patterns in our HF HCP training data, including (1) angular degradation through low-rank SH channel mixing and subsampling on an icosphere projection, (2) poor SNR/artifacting with both global/patch-wise/icosphere projection noise injection and SH channel drift, and finally (3) spatial resolution constraints through Gaussian blurring and spatial resampling down to 3-4mm resolutions. This augmentation routine yields training data that more faithfully reflects underlying brain anatomy while remaining compatible with sparse, single-shell ULF protocols. 

\vspace{0.5cm}

In our simulated spatial and angular downsampling experiment on the Connectom HCP dataset, DiffSR improved the similarity with undegraded ground-truth scans directly in SH space and showed recovery of V1 angular fidelity and signal anisotropy through FA measurements at both low spatial and angular resolutions. While these results establish that DiffSR can recover plausible anatomy under controlled resampling, they do not rule out the possibility that improvements in similarity (i.e., MAE accuracy and LNCC-based correlation) are dominated by learned population-averaged anatomic features (resulting in overfitting and hallucination of normal anatomy) rather than faithful reconstructions of subject-specific microstructure. To address this concern directly, we evaluated DiffSR in a disease type where tract-wise diffusion signal differences are distinct and well-characterized. AD and MCI provide a stringent benchmark, as disease-related microstructural changes in FA and ADC are heterogenous and confined to specific sets of limbic and association tracts \citep{Nir2013,Bozzali2002,Sexton2011,Zhang2009,Bozzali2016,Thompson2003}. Many AD/MCI-associated white matter tracts such as the fornix and uncinate fasciculus are also small and possess thin cross-sections. This creates a significant gap in analysis scope between conventional HF imaging and ULF/clinical-grade scanning systems \citep{PereraMolligodaArachchige2023,Yassa2010}, and thus an opportunity to assess the utility of DiffSR in closing this gap. The application of DiffSR to spatio-angularly downsampled DTI data with additional noise injection from control and AD/LMCI ADNI subjects showed recovery of group FA differences in key AD-related white matter tracts such as the fornix, uncinate fasciculus and inferior longitudinal fasciculus. Furthermore, quantification of group differences using Fisher linear discriminant projection revealed that DiffSR showed better discriminatory performance not only the degraded but original DTI sequence. Together, these results indicate that DiffSR preserves microstructurally meaningful diffusion contrast relevant to disease, and that DiffSR is applicable not only to ULF DTI, but more broadly to DTI acquired under constrained resolutions or noise levels. Finally, evaluation of DiffSR in our ULF/matched-HF DTI dataset provided a direct assessment of its behaviour in a real ULF sequence, where hardware constraints, low SNR, sparse angular sampling and artifacting jointly limit its fidelity. Relative to both native ULF and Beta-DSW bias correction, application of DiffSR on both native and Beta-DSW bias corrected ULF DTI substantially improved tract-wise agreement with matched HF DTI measurements for FA. Although absolute agreement for ADC remained limited across all ULF reconstructions, DiffSR, especially applied to uncorrected ULF DTI, still showed a slight improvement over Beta-DSW bias correction for ADC measurements with respect to HF. Importantly, these quantitative improvements were accompanied by qualitatively more refined streamline geometry when assessing deterministic tractography in a representative subject, particularly in thin tracts, indicating that DiffSR enhances white matter directionality in ULF rather than solely rescaling global diffusion metrics. 

\vspace{0.5cm}

While the development of a ULF DTI sequence and companion direction-dependent bias correction and superresolution methods substantially narrows the gap between HF and ULF DTI, several limitations remain that aim to motivate future hardware, sequence design and post-processing advances. Sequence-level limitations of ULF DTI remain one of the most significant and fundamental constraints. Current ULF DTI protocols \citep{Gholam2025}, including the one assessed in this study, suffer from exceptionally low SNR, coarse spatial resolutions, sparse Q-space sampling of diffusion encoding gradients, and low, single-shell diffusion sensitization (b=700$\frac{s}{mm^2}$). ULF DTI therefore severely limits the ability to segment and resolve morphology or assess the microstructure of both white and gray matter, especially in regions such as the brainstem where compact gray matter nuclei \citep{Olchanyi2025} and white matter structures \citep{Olchanyi2026} suffer from especially reduced SNR and partial voluming \citep{Sclocco2018}, or in smaller supratentorial tracts central to many neurological disorders such as the fornix in AD \citep{Bozzali2002,Sexton2011} and frontal sections of the anterior thalamic radiation in Parkinson's disease \citep{Zhang2020PD}. These limitations are further exacerbated with an extended acquisition duration (tens of minutes), which likely increases susceptibility to motion artifacting, signal drift and heat-related degradation in the absence of an active cooling system, ultimately rendering the sequence longer than what is clinically feasible. Future iterations of ULF DTI will benefit from ongoing ULF hardware advancements, including enhanced cooling management and more efficient readout strategies, which are important for stabilizing current ULF protocols \citep{Gholam2025,Natukunda2021}. Multi-spiral readout acquisitions leading to parallel k-space filling strategies have the potential to lead to significant reduction in scan time while improving acquisition stability \citep{Gholam2025}. 

\vspace{0.5cm}

Post-processing limitations related to ULF DTI also warrant significant consideration. The proposed Beta-DSW bias correction algorithm models bias artifacts as smooth, multiplicative fields and therefore does explicitly correct space-dependent rotations of the dominant diffusion direction, which commonly arise from B0 inhomogeneities and gradient nonlinearities. This represents a common source of misalignment amongst diffusion encoding gradient directions and will require additional modeling for direction-dependent distortions. In addition, the Beta and DSW priors used to fit the bias fields were derived from HCP data acquired at b-values higher than that used for our ULF DTI sequence (b=1000 and b=3000$\frac{s}{mm^2}$), which is outside of the b-value used for our ULF DTI protocol. B-value mismatch causes differences in the estimates of underlying distributions of diffusion properties such as FA and ADC \citep{Bisdas2008,Barrio-Arranz2015}, leading to difference in their representations within the atlas coefficients (e.g., both Beta and DSW rely on FA modulation) used for bias fitting and the target ULF DTI and potentially causing scale mismatches of the estimated bias coefficients. These b-value mismatches are particularly evident for ADC, which showed poor agreement with matched HF DTI for both Beta-DSW bias corrected and DiffSR-superresolved ULF DTI reconstructions. Unlike FA (which is a function of the ratio of eigenvalues recovered from the diffusion tensor estimate), ADC estimation is highly sensitive to absolute diffusion weighting and partially sequence design, and therefore differences between ULF (b=700$\frac{s}{mm^2}$, TR=700ms, TE=77.7ms, 2D EPI sequence) and matched HF (b=900$\frac{s}{mm^2}$, TR=4000ms, TE=60ms, MS-DWFSE sequence) reconstructions can result in global ADC biases. Furthermore, although we use the mean low-b to correct for global direction-invariant biases, and thus absolute diffusion during Beta-DSW bias correction, the use of only FA and V1 HCP atlas priors does not constrain absolute diffusion attenuation. These limitations are consistent with the Bland-Altman absolute biases observed in Figure $\ref{fig:Figure_7}$. In future work, the Beta-DSW model can be augmented with magnitude measurements, such as ADC, of the diffusion tensor as to regularize for large deviations in absolute diffusion.

\vspace{0.5cm}

Deviations in absolute diffusion weighting are also applicable to DiffSR. Specifically, DiffSR requires normalization of both the b0 signal and the converted SH coefficients to stabilize CNN training and inference. Inference on normalized b0 and SH channels therefore precludes the ability to learn b-value dependence on tissue contrast. In practice, we re-normalize superresolved reconstructions with scaling by the original global b0 and direction-averaged diffusion encoding gradient channel intensities. This mitigates but does not eliminate calibration differences across various acquisitions. This b-value sensitivity is evident in our synthetic resampling experiment with HCP data, where DiffSR demonstrated greater accuracy and stronger correlation with respect to un-degraded scans for SH coefficients, V1, and FA as compared to ADC. This difference in accuracy and correlation is even greater for Connectom HCP shells outside of the training dataset b-value range (b=1-3000$\frac{s}{mm^2}$), such as the b=5000$\frac{s}{mm^2}$ shell, indicating that ADC is inherently more difficult to reconstruct from normalized inputs and that the current version of DiffSR preferentially reconstructs directional and anisotropy information rather than absolute diffusivity. Finally, DiffSR was trained exclusively on HF single-shell estimates from HCP data with b-values mentioned above. Even with aggressive augmentation, there is a large domain gap with ULF DTI target acquisitions. We have nonetheless still shown that DiffSR can superresolve meaningful tissue contrast in multiple external datasets, including ULF. To further improve DiffSR, extending training to incorporate other datasets, including matched HF-ULF and multi-shell data, is key towards its generalizability. However, the former is non-trivial due to imperfect spatial alignment between different scanning sessions, and the latter because b-value-dependent signal decay demands explicit modeling over tissue compartments \citep{Tournier2019,Magin2019}.

\vspace{0.5cm}

To that nature, DiffSR was also trained and validated exclusively on single-shell sequences from HCP data b=1000$\frac{s}{mm^2}$, b=2000$\frac{s}{mm^2}$, and b=3000$\frac{s}{mm^2}$ shells), which even with our rigorous augmentation strategies results in a large domain gap between DiffSR training data and ULF DTI target data. This domain gap is even partially relevant in our AD analysis, where DiffSR reconstructions for ADC showed both a scale/variance mismatch (pointing to the above normalization/renormalization limitation) as well as attenuated within-group mean differences tracts, indicating that DiffSR potentially resorts to “blurring” (i.e., population-averaging) out-of-domain DTI. Extending training to include ULF data and incorporating loss functions in addition to angular error that directly incorporate diffusion tensor estimates such as FA and ADC is therefore an important future step. More broadly, adapting the neural network model to incorporate multi-shell DTI data is key towards furthering the generalizability of DiffSR in estimating absolute diffusion, but is non-trivial because b-value dependent signal decay demands explicit modeling over tissue compartments \citep{Tournier2019,Magin2019}. Until a multi-shell variant of DiffSR is available, the algorithm should be interpreted with caution in terms of recovering fine microstructural properties of brain tissue (which are enhanced through multi-shell acquisitions). Nonetheless, even with these limitations in mind, we have shown that DiffSR is robust across multiple DTI datasets and is thus meant to be complementary to, and not contingent on, hardware design or sequence advancements for ULF DTI. 

\vspace{0.5cm}

In summary, we introduce a ULF DTI sequence and a companion set of algorithms for its enhancement: a Q-space-aware Bayesian bias field correction algorithm coupled with an SH-based spatio-angular superresolution method, which we term DiffSR. The presented correction and superresolution algorithms consistently improve ULF diffusion metrics, tract-level analyses under severe SNR and resolution constraints, which despite current ULF acquisition limitations, are robust to acquisition differences and complement ongoing hardware and sequence-design ULF advancements. Furthermore, DiffSR improves the assessment of group-level microstructural differences in AD under similar scan degradation. Taken together, these results expand the clinical utility of ULF DTI which with ongoing improvements in hardware design, multi-shell extensions to superresolution, and multi-site validation studies will further close the HF-ULF gap for advanced DTI sequences.

\newpage

\section*{Acknowledgments}

This work was supported by NIH grants R01AG070988, 1R01EB031114, 1UM1MH130981, 1RF1AG080371, 1R21NS138995, and the American Heart Association-Tedy’s Team postdoctoral fellowship award (23POST1023166). 

\vspace{0.5cm}

Data was also provided in part by the Human Connectome Project, WU-Minn Consortium (Principal Investigators: David Van Essen and Kamil Ugurbil; 1U54MH091657) funded by the 16 NIH Institutes and Centers that support the NIH Blueprint for Neuroscience Research; and by the McDonnell Center for Systems Neuroscience at Washington University.

\vspace{0.5cm}

Data collection and sharing for Alzheimer’s disease DTI analysis in this manuscript was funded by the Alzheimer's Disease Neuroimaging Initiative (ADNI) (National Institutes of Health Grant U01 AG024904) and DOD ADNI (Department of Defense award number W81XWH-12-2-0012). ADNI is funded by the National Institute on Aging, the National Institute of Biomedical Imaging and Bioengineering, and through generous contributions from the following: AbbVie, Alzheimer’s Association; Alzheimer’s Drug Discovery Foundation; Araclon Biotech; BioClinica, Inc.; Biogen; Bristol-Myers Squibb Company; CereSpir, Inc.; Cogstate; Eisai Inc.; Elan Pharmaceuticals, Inc.; Eli Lilly and Company; EuroImmun; F. Hoffmann-La Roche Ltd and its affiliated company Genentech, Inc.; Fujirebio; GE Healthcare; IXICO Ltd.; Janssen Alzheimer Immunotherapy Research \& Development, LLC.; Johnson \& Johnson Pharmaceutical Research \& Development LLC.; Lumosity; Lundbeck; Merck \& Co., Inc.; Meso Scale Diagnostics, LLC.; NeuroRx Research; Neurotrack Technologies; Novartis Pharmaceuticals Corporation; Pfizer Inc.; Piramal Imaging; Servier; Takeda Pharmaceutical Company; and Transition Therapeutics. The Canadian Institutes of Health Research is providing funds to support ADNI clinical sites in Canada. Private sector contributions are facilitated by the Foundation for the National Institutes of Health (www.fnih.org). The grantee organization is the Northern California Institute for Research and Education, and the study is coordinated by the Alzheimer’s Therapeutic Research Institute at the University of Southern California. ADNI data are disseminated by the Laboratory for Neuro Imaging at the University of Southern California.

\vspace{0.5cm}

Matthew S. Rosen acknowledges the generous support of the Kiyomi and Ed Baird MGH Scholar award. MSR is a founder and equity holder of Hyperfine, Inc. MSR is a consultant and equity holder of DeepSpin GmbH.

\vspace{0.5cm}

We are grateful for the feedback provided by Dr. Iman Aganj regarding spherical harmonic operations.

\bibliographystyle{unsrtnat} 
\bibliography{DiffSR_references}

@article{Olchanyi2026,
   author = {Mark D. Olchanyi and David R. Schreier and Jian Li and Chiara Maffei and Annabel Sorby-Adams and Hannah C. Kinney and Brian C. Healy and Holly J. Freeman and Jared Shless and Christophe Destrieux and Henry Tregidgo and Juan Eugenio Iglesias and Emery N. Brown and Brian L. Edlow},
   doi = {10.1073/pnas.2509321123},
   issn = {0027-8424},
   issue = {6},
   journal = {Proceedings of the National Academy of Sciences},
   month = {2},
   title = {Probabilistic mapping and automated segmentation of human brainstem white matter bundles},
   volume = {123},
   year = {2026}
}

@article{Oishi2014,
   author = {Kenichi Oishi and Constantine G. Lyketsos},
   doi = {10.3389/fnagi.2014.00241},
   issn = {1663-4365},
   journal = {Frontiers in Aging Neuroscience},
   month = {9},
   title = {Alzheimer's disease and the fornix},
   volume = {6},
   year = {2014}
}

@article{Janelle2022,
   author = {Felix Janelle and Christian Iorio-Morin and Sabrina D'amour and David Fortin},
   doi = {10.3389/fneur.2022.794618},
   issn = {1664-2295},
   journal = {Frontiers in Neurology},
   month = {4},
   title = {Superior Longitudinal Fasciculus: A Review of the Anatomical Descriptions With Functional Correlates},
   volume = {13},
   year = {2022}
}

@article{DeLuca2022,
   author = {Alberto De Luca and Suheyla Cetin Karayumak and Alexander Leemans and Yogesh Rathi and Stephan Swinnen and Jolien Gooijers and Amanda Clauwaert and Roald Bahr and Stian Bahr Sandmo and Nir Sochen and David Kaufmann and Marc Muehlmann and Geert-Jan Biessels and Inga Koerte and Ofer Pasternak},
   doi = {10.1016/j.neuroimage.2022.119439},
   issn = {10538119},
   journal = {NeuroImage},
   month = {10},
   pages = {119439},
   title = {Cross-site harmonization of multi-shell diffusion MRI measures based on rotational invariant spherical harmonics (RISH)},
   volume = {259},
   year = {2022}
}

@article{Chen2025,
   author = {Yunwei Chen and Jialong Li and Qiqi Lu and Ye Wu and Xiaoming Liu and Yuanyuan Gao and Yanqiu Feng and Zhicheng Zhang and Xinyuan Zhang},
   doi = {10.1109/JBHI.2024.3471769},
   issn = {2168-2194},
   issue = {1},
   journal = {IEEE Journal of Biomedical and Health Informatics},
   month = {1},
   pages = {456-467},
   title = {Spherical Harmonics-Based Deep Learning Achieves Generalized and Accurate Diffusion Tensor Imaging},
   volume = {29},
   year = {2025}
}

@inproceedings{Tobin2017,
   author = {Josh Tobin and Rachel Fong and Alex Ray and Jonas Schneider and Wojciech Zaremba and Pieter Abbeel},
   doi = {10.1109/IROS.2017.8202133},
   isbn = {978-1-5386-2682-5},
   booktitle = {2017 IEEE/RSJ International Conference on Intelligent Robots and Systems (IROS)},
   month = {9},
   pages = {23-30},
   publisher = {IEEE},
   title = {Domain randomization for transferring deep neural networks from simulation to the real world},
   year = {2017}
}

@article{Bisdas2008,
   author = {S. Bisdas and D.E. Bohning and N. Bešenski and J.S. Nicholas and Z. Rumboldt},
   doi = {10.3174/ajnr.A1044},
   issn = {0195-6108},
   issue = {6},
   journal = {American Journal of Neuroradiology},
   month = {6},
   pages = {1128-1133},
   title = {Reproducibility, Interrater Agreement, and Age-Related Changes of Fractional Anisotropy Measures at 3T in Healthy Subjects: Effect of the Applied b-Value},
   volume = {29},
   year = {2008}
}

@article{Barrio-Arranz2015,
   author = {Gonzalo Barrio-Arranz and Rodrigo de Luis-García and Antonio Tristán-Vega and Marcos Martín-Fernández and Santiago Aja-Fernández},
   doi = {10.1371/journal.pone.0137905},
   issn = {1932-6203},
   issue = {10},
   journal = {PLOS ONE},
   month = {10},
   pages = {e0137905},
   title = {Impact of MR Acquisition Parameters on DTI Scalar Indexes: A Tractography Based Approach},
   volume = {10},
   year = {2015}
}

@article{Natukunda2021,
   author = {Faith Natukunda and Theodora M. Twongyirwe and Steven J. Schiff and Johnes Obungoloch},
   doi = {10.1186/s42490-021-00048-6},
   issn = {2524-4426},
   issue = {1},
   journal = {BMC Biomedical Engineering},
   month = {12},
   pages = {3},
   title = {Approaches in cooling of resistive coil-based low-field Magnetic Resonance Imaging (MRI) systems for application in low resource settings},
   volume = {3},
   year = {2021}
}

@article{Gopinath2025,
   author = {Karthik Gopinath and Douglas N. Greve and Colin Magdamo and Steve Arnold and Sudeshna Das and Oula Puonti and Juan Eugenio Iglesias},
   doi = {10.1016/j.media.2025.103608},
   issn = {13618415},
   journal = {Medical Image Analysis},
   month = {7},
   pages = {103608},
   title = {“Recon-all-clinical”: Cortical surface reconstruction and analysis of heterogeneous clinical brain MRI},
   volume = {103},
   year = {2025}
}

@article{Liu2025,
   author = {Peirong Liu and Oula Puonti and Xiaoling Hu and Karthik Gopinath and Annabel Sorby-Adams and Daniel Alexander and W. Taylor Kimberly and Juan E. Iglesias},
   journal = {arXiv},
   title = {A Modality-agnostic Multi-task Foundation Model for Human Brain Imaging},
   year = {2025}
}

@article{Setsompop2013,
   author = {K. Setsompop and R. Kimmlingen and E. Eberlein and T. Witzel and J. Cohen-Adad and J.A. McNab and B. Keil and M.D. Tisdall and P. Hoecht and P. Dietz and S.F. Cauley and V. Tountcheva and V. Matschl and V.H. Lenz and K. Heberlein and A. Potthast and H. Thein and J. Van Horn and A. Toga and F. Schmitt and D. Lehne and B.R. Rosen and V. Wedeen and L.L. Wald},
   doi = {10.1016/j.neuroimage.2013.05.078},
   issn = {10538119},
   journal = {NeuroImage},
   month = {10},
   pages = {220-233},
   title = {Pushing the limits of in vivo diffusion MRI for the Human Connectome Project},
   volume = {80},
   year = {2013}
}

@article{Wieczorek2018,
   author = {Mark A. Wieczorek and Matthias Meschede},
   doi = {10.1029/2018GC007529},
   issn = {1525-2027},
   issue = {8},
   journal = {Geochemistry, Geophysics, Geosystems},
   month = {8},
   pages = {2574-2592},
   title = {SHTools: Tools for Working with Spherical Harmonics},
   volume = {19},
   year = {2018}
}

@article{Avants2011,
   author = {Brian B. Avants and Nicholas J. Tustison and Gang Song and Philip A. Cook and Arno Klein and James C. Gee},
   doi = {10.1016/j.neuroimage.2010.09.025},
   issn = {10538119},
   issue = {3},
   journal = {NeuroImage},
   month = {2},
   pages = {2033-2044},
   title = {A reproducible evaluation of ANTs similarity metric performance in brain image registration},
   volume = {54},
   year = {2011}
}

@article{ZHANG2006,
   author = {H ZHANG and P YUSHKEVICH and D ALEXANDER and J GEE},
   doi = {10.1016/j.media.2006.06.004},
   issn = {13618415},
   issue = {5},
   journal = {Medical Image Analysis},
   month = {10},
   pages = {764-785},
   title = {Deformable registration of diffusion tensor MR images with explicit orientation optimization},
   volume = {10},
   year = {2006}
}

@article{Sorby-Adams2024,
   author = {Annabel J. Sorby-Adams and Jennifer Guo and Pablo Laso and John E. Kirsch and Julia Zabinska and Ana-Lucia Garcia Guarniz and Pamela W. Schaefer and Seyedmehdi Payabvash and Adam de Havenon and Matthew S. Rosen and Kevin N. Sheth and Teresa Gomez-Isla and J. Eugenio Iglesias and W. Taylor Kimberly},
   doi = {10.1038/s41467-024-54972-x},
   issn = {2041-1723},
   issue = {1},
   journal = {Nature Communications},
   month = {12},
   pages = {10488},
   title = {Portable, low-field magnetic resonance imaging for evaluation of Alzheimer’s disease},
   volume = {15},
   year = {2024}
}

@article{Farnan2025,
   author = {Ava Farnan and Annabel Sorby‐Adams and Jennifer Guo and Pablo Laso and Amanda Desenna and John Kirsch and Julia Zabinska and John R Dickson and Liliana A Ramirez Gomez and Pamela Shaefer and Seyedmehdi Payabvash and Adam de Havenon and Mattew Rosen and Kevin Sheth and Juan Eugenio Iglesias and Teresa Gomez‐Isla and W. Taylor Kimberly},
   doi = {10.1002/alz70856_103075},
   issn = {1552-5260},
   issue = {S2},
   journal = {Alzheimer's \& Dementia},
   month = {12},
   title = {Portable, Low‐Field MRI for Alzheimer's Disease: Detecting Patterns of Atrophy Using Machine Learning Pipelines},
   volume = {21},
   year = {2025}
}

@article{Olchanyi2025,
   author = {Mark D. Olchanyi and Jean Augustinack and Robin L. Haynes and Laura D. Lewis and Nicholas Cicero and Jian Li and Christophe Destrieux and Rebecca D. Folkerth and Hannah C. Kinney and Bruce Fischl and Emery N. Brown and Juan Eugenio Iglesias and Brian L. Edlow},
   doi = {10.1002/hbm.70357},
   issn = {1065-9471},
   issue = {14},
   journal = {Human Brain Mapping},
   month = {10},
   title = {Automated MRI Segmentation of Brainstem Nuclei Critical to Consciousness},
   volume = {46},
   year = {2025}
}

@inproceedings{Iglesias2019,
   author = {Juan Eugenio Iglesias and Koen Van Leemput and Polina Golland and Anastasia Yendiki},
   doi = {10.1007/978-3-030-20351-1_60},
   booktitle = {Inf Process Med Imaging},
   pages = {767-779},
   title = {Joint Inference on Structural and Diffusion MRI for Sequence-Adaptive Bayesian Segmentation of Thalamic Nuclei with Probabilistic Atlases},
   year = {2019}
}

@article{VanLeemput1999,
   author = {K. Van Leemput and F. Maes and D. Vandermeulen and P. Suetens},
   doi = {10.1109/42.811268},
   issn = {02780062},
   issue = {10},
   journal = {IEEE Transactions on Medical Imaging},
   pages = {885-896},
   title = {Automated model-based bias field correction of MR images of the brain},
   volume = {18},
   year = {1999}
}

@inbook{Chen2022,
   author = {Geng Chen and Haotian Jiang and Jiannan Liu and Jiquan Ma and Hui Cui and Yong Xia and Pew-Thian Yap},
   doi = {10.1007/978-3-031-16431-6_11},
   pages = {113-122},
   journal = {Medical Image Computing and Computer Assisted Intervention},
   publisher = {Springer Nature},
   title = {Hybrid Graph Transformer for Tissue Microstructure Estimation with Undersampled Diffusion MRI Data},
   year = {2022}
}

@inbook{Chen2020,
   author = {Geng Chen and Yoonmi Hong and Yongqin Zhang and Jaeil Kim and Khoi Minh Huynh and Jiquan Ma and Weili Lin and Dinggang Shen and Pew-Thian Yap},
   doi = {10.1007/978-3-030-59728-3_28},
   journal = {Medical Image Computing and Computer Assisted Intervention},
   publisher = {Springer Nature},
   pages = {280-290},
   title = {Estimating Tissue Microstructure with Undersampled Diffusion Data via Graph Convolutional Neural Networks},
   year = {2020}
}

@inproceedings{Nath2020,
   author = {Vishwesh Nath and Sudhir K. Pathak and Kurt G. Schilling and Walter Schneider and Bennett A. Landman},
   doi = {10.1117/12.2549455},
   editor = {Bennett A. Landman and Ivana Išgum},
   isbn = {9781510633933},
   booktitle = {Medical Imaging 2020: Image Processing},
   month = {3},
   pages = {27},
   publisher = {SPIE},
   title = {Deep learning estimation of multi-tissue constrained spherical deconvolution with limited single shell DW-MRI},
   year = {2020}
}

@article{Rathi2014,
   author = {Y. Rathi and O. Michailovich and F. Laun and K. Setsompop and P.E. Grant and C.-F. Westin},
   doi = {10.1016/j.media.2014.06.003},
   issn = {13618415},
   issue = {7},
   journal = {Medical Image Analysis},
   month = {10},
   pages = {1143-1156},
   title = {Multi-shell diffusion signal recovery from sparse measurements},
   volume = {18},
   year = {2014}
}

@article{Ordinola2025,
   author = {Alfredo Ordinola and David Abramian and Magnus Herberthson and Anders Eklund and Evren Özarslan},
   doi = {10.1038/s41598-025-90972-7},
   issn = {2045-2322},
   issue = {1},
   journal = {Scientific Reports},
   month = {2},
   pages = {6580},
   title = {Super-resolution mapping of anisotropic tissue structure with diffusion MRI and deep learning},
   volume = {15},
   year = {2025}
}

@article{Ewert2024,
   author = {Christian Ewert and David Kügler and Rüdiger Stirnberg and Alexandra Koch and Anastasia Yendiki and Martin Reuter},
   doi = {10.1162/imag_a_00121},
   issn = {2837-6056},
   journal = {Imaging Neuroscience},
   month = {4},
   title = {Geometric deep learning for diffusion MRI signal reconstruction with continuous samplings (DISCUS)},
   volume = {2},
   year = {2024}
}

@article{LeBihan2001,
   author = {Denis Le Bihan and Jean‐François Mangin and Cyril Poupon and Chris A. Clark and Sabina Pappata and Nicolas Molko and Hughes Chabriat},
   doi = {10.1002/jmri.1076},
   issn = {1053-1807},
   issue = {4},
   journal = {Journal of Magnetic Resonance Imaging},
   month = {4},
   pages = {534-546},
   title = {Diffusion tensor imaging: Concepts and applications},
   volume = {13},
   year = {2001}
}

@article{Calabrese2016,
   author = {Evan Calabrese},
   doi = {10.3389/fnana.2016.00045},
   issn = {1662-5129},
   journal = {Frontiers in Neuroanatomy},
   month = {5},
   title = {Diffusion Tractography in Deep Brain Stimulation Surgery: A Review},
   volume = {10},
   year = {2016}
}

@article{Magin2019,
   author = {Richard L. Magin and M. Muge Karaman and Matt G. Hall and Wenzhen Zhu and Xiaohong Joe Zhou},
   doi = {10.1016/j.mri.2018.09.034},
   issn = {0730725X},
   journal = {Magnetic Resonance Imaging},
   month = {2},
   pages = {110-118},
   title = {Capturing complexity of the diffusion-weighted MR signal decay},
   volume = {56},
   year = {2019}
}

@article{Michailovich2011,
   author = {Oleg Michailovich and Yogesh Rathi and Sudipto Dolui},
   doi = {10.1109/TMI.2011.2142189},
   issn = {0278-0062},
   issue = {5},
   journal = {IEEE Transactions on Medical Imaging},
   month = {5},
   pages = {1100-1115},
   title = {Spatially Regularized Compressed Sensing for High Angular Resolution Diffusion Imaging},
   volume = {30},
   year = {2011}
}

@article{Coup2013,
   author = {Pierrick Coupé and José V. Manjón and Maxime Chamberland and Maxime Descoteaux and Bassem Hiba},
   doi = {10.1016/j.neuroimage.2013.06.030},
   issn = {10538119},
   journal = {NeuroImage},
   month = {12},
   pages = {245-261},
   title = {Collaborative patch-based super-resolution for diffusion-weighted images},
   volume = {83},
   year = {2013}
}

@article{Maffei2021,
   author = {C. Maffei and C. Lee and M. Planich and M. Ramprasad and N. Ravi and D. Trainor and Z. Urban and M. Kim and R.J. Jones and A. Henin and S.G. Hofmann and D.A. Pizzagalli and R.P. Auerbach and J.D.E. Gabrieli and S. Whitfield-Gabrieli and D.N. Greve and S.N. Haber and A. Yendiki},
   doi = {10.1016/j.neuroimage.2021.118706},
   issn = {10538119},
   journal = {NeuroImage},
   month = {12},
   pages = {118706},
   title = {Using diffusion MRI data acquired with ultra-high gradient strength to improve tractography in routine-quality data},
   volume = {245},
   year = {2021}
}

@article{Winter2021,
   author = {Mia Winter and Emma C Tallantyre and Thomas A W Brice and Neil P Robertson and Derek K Jones and Maxime Chamberland},
   doi = {10.1093/braincomms/fcab065},
   issn = {2632-1297},
   issue = {2},
   journal = {Brain Communications},
   month = {4},
   title = {Tract-specific MRI measures explain learning and recall differences in multiple sclerosis},
   volume = {3},
   year = {2021}
}

@article{Filippi2019,
   author = {Massimo Filippi and Paolo Preziosa and Brenda L Banwell and Frederik Barkhof and Olga Ciccarelli and Nicola De Stefano and Jeroen J G Geurts and Friedemann Paul and Daniel S Reich and Ahmed T Toosy and Anthony Traboulsee and Mike P Wattjes and Tarek A Yousry and Achim Gass and Catherine Lubetzki and Brian G Weinshenker and Maria A Rocca},
   doi = {10.1093/brain/awz144},
   issn = {0006-8950},
   issue = {7},
   journal = {Brain},
   month = {7},
   pages = {1858-1875},
   title = {Assessment of lesions on magnetic resonance imaging in multiple sclerosis: practical guidelines},
   volume = {142},
   year = {2019}
}

@article{Yassa2010,
   author = {Michael A. Yassa and L. Tugan Muftuler and Craig E. L. Stark},
   doi = {10.1073/pnas.1002113107},
   issn = {0027-8424},
   issue = {28},
   journal = {Proceedings of the National Academy of Sciences},
   month = {7},
   pages = {12687-12691},
   title = {Ultrahigh-resolution microstructural diffusion tensor imaging reveals perforant path degradation in aged humans in vivo},
   volume = {107},
   year = {2010}
}

@article{PereraMolligodaArachchige2023,
   author = {Arosh S. Perera Molligoda Arachchige and Anton Kristoffer Garner},
   doi = {10.3934/Neuroscience.2023030},
   issn = {2373-7972},
   issue = {4},
   journal = {AIMS Neuroscience},
   pages = {401-422},
   title = {Seven Tesla MRI in Alzheimer's disease research: State of the art and future directions: A narrative review},
   volume = {10},
   year = {2023}
}

@article{Gholam2025,
   author = {James Gholam and Phil Schmid and Joshua Ametepe and Alix Plumley and Leandro Beltrachini and Francesco Padormo and Rui Teixeira and Rafael OHalloran and Kaloian Petkov and Klaus Engel and Steven CR Williams and Sean Deoni and Mara Cercignani and Derek K Jones},
   journal = {arXiv},
   month = {6},
   title = {Diffusion Tensor MRI and Spherical-Deconvolution-Based Tractography on an Ultra-Low Field Portable MRI System},
   year = {2025}
}

@article{Stejskal1965,
   author = {E. O. Stejskal and J. E. Tanner},
   doi = {10.1063/1.1695690},
   issn = {0021-9606},
   issue = {1},
   journal = {The Journal of Chemical Physics},
   month = {1},
   pages = {288-292},
   title = {Spin Diffusion Measurements: Spin Echoes in the Presence of a Time-Dependent Field Gradient},
   volume = {42},
   year = {1965}
}

@article{Cooley2020,
   author = {Clarissa Z. Cooley and Patrick C. McDaniel and Jason P. Stockmann and Sai Abitha Srinivas and Stephen F. Cauley and Monika Śliwiak and Charlotte R. Sappo and Christopher F. Vaughn and Bastien Guerin and Matthew S. Rosen and Michael H. Lev and Lawrence L. Wald},
   doi = {10.1038/s41551-020-00641-5},
   issn = {2157-846X},
   issue = {3},
   journal = {Nature Biomedical Engineering},
   month = {11},
   pages = {229-239},
   title = {A portable scanner for magnetic resonance imaging of the brain},
   volume = {5},
   year = {2020}
}

@article{Liu2021,
   author = {Yilong Liu and Alex T. L. Leong and Yujiao Zhao and Linfang Xiao and Henry K. F. Mak and Anderson Chun On Tsang and Gary K. K. Lau and Gilberto K. K. Leung and Ed X. Wu},
   doi = {10.1038/s41467-021-27317-1},
   issn = {2041-1723},
   issue = {1},
   journal = {Nature Communications},
   month = {12},
   pages = {7238},
   title = {A low-cost and shielding-free ultra-low-field brain MRI scanner},
   volume = {12},
   year = {2021}
}

@article{Homeier1996,
   author = {Herbert H.H. Homeier and E.Otto Steinborn},
   doi = {10.1016/S0166-1280(96)90531-X},
   issn = {01661280},
   journal = {Journal of Molecular Structure: THEOCHEM},
   month = {9},
   pages = {31-37},
   title = {Some properties of the coupling coefficients of real spherical harmonics and their relation to Gaunt coefficients},
   volume = {368},
   year = {1996}
}

@article{Politis2024,
   author = {Archontis Politis},
   journal = {arXiv},
   title = {Gaunt coefficients for complex and real spherical harmonics with applications to spherical array processing and Ambisonics},
   year = {2024}
}

@article{Ametepe2024,
   author = {Joshua Mawuli Ametepe and James Gholam and Leandro Beltrachini and Mara Cercignani and Derek Kenton Jones},
   doi = {10.1101/2024.08.19.24312228},
   journal = {medRxiv},
   month = {8},
   title = {Machine-Learning Enhanced Diffusion Tensor Imaging with Four Encoding Directions},
   year = {2024}
}

@article{Zeng2022,
   author = {Rui Zeng and Jinglei Lv and He Wang and Luping Zhou and Michael Barnett and Fernando Calamante and Chenyu Wang},
   doi = {10.1016/j.media.2022.102431},
   issn = {13618415},
   journal = {Medical Image Analysis},
   month = {7},
   pages = {102431},
   title = {FOD-Net: A deep learning method for fiber orientation distribution angular super resolution},
   volume = {79},
   year = {2022}
}

@article{Lyon2022,
   author = {Matthew Lyon and Paul Armitage and Mauricio Álvarez},
   journal = {arXiv},
   title = {Angular Super-Resolution in Diffusion MRI with a 3D Recurrent Convolutional Autoencoder},
   year = {2022}
}

@article{vandeWiele2001,
   author = {J. van de Wiele},
   journal = {Annales de Physique},
   pages = {1-169},
   title = {Rotations et Moments Angulaires en Mecanique Quantique - Rotations and angular moments in quantum mechanics},
   volume = {26},
   year = {2001}
}

@article{Bourke2021,
   author = {Niall J Bourke and Maria Yanez Lopez and Peter O Jenkins and Sara De Simoni and James H Cole and Pete Lally and Emma-Jane Mallas and Hui Zhang and David J Sharp},
   doi = {10.1093/braincomms/fcab006},
   issn = {2632-1297},
   issue = {2},
   journal = {Brain Communications},
   month = {4},
   title = {Traumatic brain injury: a comparison of diffusion and volumetric magnetic resonance imaging measures},
   volume = {3},
   year = {2021}
}

@article{Petersen2010,
   author = {R. C. Petersen and P. S. Aisen and L. A. Beckett and M. C. Donohue and A. C. Gamst and D. J. Harvey and C. R. Jack and W. J. Jagust and L. M. Shaw and A. W. Toga and J. Q. Trojanowski and M. W. Weiner},
   doi = {10.1212/WNL.0b013e3181cb3e25},
   issn = {0028-3878},
   issue = {3},
   journal = {Neurology},
   month = {1},
   pages = {201-209},
   title = {Alzheimer's Disease Neuroimaging Initiative (ADNI)},
   volume = {74},
   year = {2010}
}

@article{Zhang2020PD,
   author = {Yu Zhang and Marc A. Burock},
   doi = {10.3389/fneur.2020.531993},
   issn = {1664-2295},
   journal = {Frontiers in Neurology},
   month = {9},
   title = {Diffusion Tensor Imaging in Parkinson's Disease and Parkinsonian Syndrome: A Systematic Review},
   volume = {11},
   year = {2020}
}

@article{Bozzali2002,
   author = {M Bozzali},
   doi = {10.1136/jnnp.72.6.742},
   issn = {00223050},
   issue = {6},
   journal = {Journal of Neurology, Neurosurgery \& Psychiatry},
   month = {6},
   pages = {742-746},
   title = {White matter damage in Alzheimer's disease assessed in vivo using diffusion tensor magnetic resonance imaging},
   volume = {72},
   year = {2002}
}

@article{Tournier2019,
   author = {J-Donald Tournier and Robert Smith and David Raffelt and Rami Tabbara and Thijs Dhollander and Maximilian Pietsch and Daan Christiaens and Ben Jeurissen and Chun-Hung Yeh and Alan Connelly},
   doi = {10.1016/j.neuroimage.2019.116137},
   issn = {10538119},
   journal = {NeuroImage},
   month = {11},
   pages = {116137},
   title = {MRtrix3: A fast, flexible and open software framework for medical image processing and visualisation},
   volume = {202},
   year = {2019}
}

@article{Sotiropoulos2013,
   author = {S. N. Sotiropoulos and S. Moeller and S. Jbabdi and J. Xu and J. L. Andersson and E. J. Auerbach and E. Yacoub and D. Feinberg and K. Setsompop and L. L. Wald and T. E. J. Behrens and K. Ugurbil and C. Lenglet},
   doi = {10.1002/mrm.24623},
   issn = {0740-3194},
   issue = {6},
   journal = {Magnetic Resonance in Medicine},
   month = {12},
   pages = {1682-1689},
   title = {Effects of image reconstruction on fiber orientation mapping from multichannel diffusion MRI: Reducing the noise floor using SENSE},
   volume = {70},
   year = {2013}
}

@article{Kinnunen2011,
   author = {K. M. Kinnunen and R. Greenwood and J. H. Powell and R. Leech and P. C. Hawkins and V. Bonnelle and M. C. Patel and S. J. Counsell and D. J. Sharp},
   doi = {10.1093/brain/awq347},
   issn = {0006-8950},
   issue = {2},
   journal = {Brain},
   month = {2},
   pages = {449-463},
   title = {White matter damage and cognitive impairment after traumatic brain injury},
   volume = {134},
   year = {2011}
}

@article{Sheth2021,
   author = {Kevin N. Sheth and Mercy H. Mazurek and Matthew M. Yuen and Bradley A. Cahn and Jill T. Shah and Adrienne Ward and Jennifer A. Kim and Emily J. Gilmore and Guido J. Falcone and Nils Petersen and Kevin T. Gobeske and Firas Kaddouh and David Y. Hwang and Joseph Schindler and Lauren Sansing and Charles Matouk and Jonathan Rothberg and Gordon Sze and Jonathan Siner and Matthew S. Rosen and Serena Spudich and W. Taylor Kimberly},
   doi = {10.1001/jamaneurol.2020.3263},
   issn = {2168-6149},
   issue = {1},
   journal = {JAMA Neurology},
   month = {1},
   pages = {41},
   title = {Assessment of Brain Injury Using Portable, Low-Field Magnetic Resonance Imaging at the Bedside of Critically Ill Patients},
   volume = {78},
   year = {2021}
}

@article{Bozzali2016,
   author = {Marco Bozzali and Laura Serra and Mara Cercignani},
   doi = {10.1097/WCO.0000000000000345},
   issn = {1350-7540},
   issue = {4},
   journal = {Current Opinion in Neurology},
   month = {8},
   pages = {437-444},
   title = {Quantitative MRI to understand Alzheimer's disease pathophysiology},
   volume = {29},
   year = {2016}
}

@article{Smith2004,
   author = {Stephen M Smith and Mark Jenkinson and Mark W Woolrich and Christian F Beckmann and Timothy E J Behrens and Heidi Johansen-Berg and Peter R Bannister and Marilena De Luca and Ivana Drobnjak and David E Flitney and Rami K Niazy and James Saunders and John Vickers and Yongyue Zhang and Nicola De Stefano and J Michael Brady and Paul M Matthews},
   doi = {10.1016/j.neuroimage.2004.07.051},
   issn = {1053-8119},
   journal = {NeuroImage},
   pages = {S208-S219},
   publisher = {Elsevier BV},
   title = {Advances in functional and structural MR image analysis and implementation as FSL},
   volume = {23},
   url = {http://dx.doi.org/10.1016/j.neuroimage.2004.07.051},
   year = {2004}
}

@article{Sexton2011,
   author = {Claire E. Sexton and Ukwuori G. Kalu and Nicola Filippini and Clare E. Mackay and Klaus P. Ebmeier},
   doi = {10.1016/j.neurobiolaging.2010.05.019},
   issn = {01974580},
   issue = {12},
   journal = {Neurobiology of Aging},
   month = {12},
   pages = {2322.e5-2322.e18},
   title = {A meta-analysis of diffusion tensor imaging in mild cognitive impairment and Alzheimer's disease},
   volume = {32},
   year = {2011}
}

@article{Ronneberger2015,
   author = {Olaf Ronneberger and Philipp Fischer and Thomas Brox},
   month = {5},
   title = {U-Net: Convolutional Networks for Biomedical Image Segmentation},
   journal = {arXiv},
   publisher = {arXiv},
   doi = {10.48550/arXiv.1505.04597},
   year = {2015}
}

@article{Zhang2009,
   author = {Y. Zhang and N. Schuff and A.-T. Du and H. J. Rosen and J. H. Kramer and M. L. Gorno-Tempini and B. L. Miller and M. W. Weiner},
   doi = {10.1093/brain/awp071},
   issn = {0006-8950},
   issue = {9},
   journal = {Brain},
   month = {9},
   pages = {2579-2592},
   title = {White matter damage in frontotemporal dementia and Alzheimer's disease measured by diffusion MRI},
   volume = {132},
   year = {2009}
}

@article{Nir2013,
   author = {Talia M. Nir and Neda Jahanshad and Julio E. Villalon-Reina and Arthur W. Toga and Clifford R. Jack and Michael W. Weiner and Paul M. Thompson},
   doi = {10.1016/j.nicl.2013.07.006},
   issn = {22131582},
   journal = {NeuroImage: Clinical},
   pages = {180-195},
   title = {Effectiveness of regional DTI measures in distinguishing Alzheimer's disease, MCI, and normal aging},
   volume = {3},
   year = {2013}
}

@article{Thompson2003,
   author = {Paul M. Thompson and Kiralee M. Hayashi and Greig de Zubicaray and Andrew L. Janke and Stephen E. Rose and James Semple and David Herman and Michael S. Hong and Stephanie S. Dittmer and David M. Doddrell and Arthur W. Toga},
   doi = {10.1523/JNEUROSCI.23-03-00994.2003},
   issn = {0270-6474},
   issue = {3},
   journal = {The Journal of Neuroscience},
   month = {2},
   pages = {994-1005},
   title = {Dynamics of Gray Matter Loss in Alzheimer's Disease},
   volume = {23},
   year = {2003}
}

@article{Benjamini1995,
   author = {Yoav Benjamini and Yosef Hochberg},
   doi = {10.1111/j.2517-6161.1995.tb02031.x},
   issn = {1369-7412},
   issue = {1},
   journal = {Journal of the Royal Statistical Society Series B: Statistical Methodology},
   month = {1},
   pages = {289-300},
   title = {Controlling the False Discovery Rate: A Practical and Powerful Approach to Multiple Testing},
   volume = {57},
   year = {1995}
}

@inproceedings{Kingma2014,
   author = {Diederik Kingma and Jimmy Ba},
   booktitle = {Adam: A Method for Stochastic Optimization},
   publisher = {arXiv},
   title = {Adam: A Method for Stochastic Optimization},
   year = {2014}
}

@article{Yendiki2011,
   author = {Anastasia Yendiki and Patricia Panneck and Priti Srinivasan and Allison Stevens and Lilla Zöllei and Jean Augustinack and Ruopeng Wang and David Salat and Stefan Ehrlich and Tim Behrens and Saad Jbabdi and Randy Gollub and Bruce Fischl},
   city = {Switzerland},
   doi = {10.3389/fninf.2011.00023},
   issn = {1662-5196},
   journal = {Frontiers in neuroinformatics},
   month = {10},
   pages = {23},
   pmid = {22016733},
   title = {Automated probabilistic reconstruction of white-matter pathways in health and disease using an atlas of the underlying anatomy},
   volume = {5},
   url = {https://pubmed.ncbi.nlm.nih.gov/22016733 https://www.ncbi.nlm.nih.gov/pmc/articles/PMC3193073/},
   year = {2011}
}

@article{Weiner2017,
   author = {Michael W Weiner and Dallas P Veitch and Paul S Aisen and Laurel A Beckett and Nigel J Cairns and Robert C Green and Danielle Harvey and Clifford R Jack Jr and William Jagust and John C Morris and Ronald C Petersen and Jennifer Salazar and Andrew J Saykin and Leslie M Shaw and Arthur W Toga and John Q Trojanowski and Alzheimer's Disease Neuroimaging Initiative},
   city = {United States},
   doi = {10.1016/j.jalz.2016.10.006},
   edition = {2016/12/05},
   issn = {1552-5279},
   issue = {5},
   journal = {Alzheimer's \& dementia : the journal of the Alzheimer's Association},
   month = {5},
   pages = {561-571},
   pmid = {27931796},
   title = {The Alzheimer's Disease Neuroimaging Initiative 3: Continued innovation for clinical trial improvement},
   volume = {13},
   url = {https://pubmed.ncbi.nlm.nih.gov/27931796 https://www.ncbi.nlm.nih.gov/pmc/articles/PMC5536850/},
   year = {2017}
}

@article{Tregidgo2023BAYES,
   author = {Henry F J Tregidgo and Sonja Soskic and Juri Althonayan and Chiara Maffei and Koen Van Leemput and Polina Golland and Ricardo Insausti and Garikoitz Lerma-Usabiaga and César Caballero-Gaudes and Pedro M Paz-Alonso and Anastasia Yendiki and Daniel C Alexander and Martina Bocchetta and Jonathan D Rohrer and Juan Eugenio Iglesias and Alzheimer's Disease Neuroimaging Initiative},
   city = {United States},
   doi = {10.1016/j.neuroimage.2023.120129},
   edition = {2023/04/22},
   issn = {1095-9572},
   journal = {NeuroImage},
   month = {7},
   pages = {120129},
   pmid = {37088323},
   title = {Accurate Bayesian segmentation of thalamic nuclei using diffusion MRI and an improved histological atlas},
   volume = {274},
   url = {https://pubmed.ncbi.nlm.nih.gov/37088323 https://www.ncbi.nlm.nih.gov/pmc/articles/PMC10636587/},
   year = {2023}
}

@article{Sclocco2018,
   author = {Roberta Sclocco and Florian Beissner and Marta Bianciardi and Jonathan R Polimeni and Vitaly Napadow},
   city = {United States},
   doi = {10.1016/j.neuroimage.2017.02.052},
   edition = {2017/02/21},
   issn = {1095-9572},
   journal = {NeuroImage},
   month = {3},
   pages = {412-426},
   pmid = {28232189},
   title = {Challenges and opportunities for brainstem neuroimaging with ultrahigh field MRI},
   volume = {168},
   url = {https://pubmed.ncbi.nlm.nih.gov/28232189 https://www.ncbi.nlm.nih.gov/pmc/articles/PMC5777900/},
   year = {2018}
}

@article{Jenkinson2012,
   author = {Mark Jenkinson and Christian F Beckmann and Timothy E J Behrens and Mark W Woolrich and Stephen M Smith},
   doi = {10.1016/j.neuroimage.2011.09.015},
   issn = {1053-8119},
   issue = {2},
   journal = {NeuroImage},
   pages = {782-790},
   publisher = {Elsevier BV},
   title = {FSL},
   volume = {62},
   url = {http://dx.doi.org/10.1016/j.neuroimage.2011.09.015},
   year = {2012}
}

@article{Iglesias2023,
   author = {Juan E Iglesias and Benjamin Billot and Yaël Balbastre and Colin Magdamo and Steven E Arnold and Sudeshna Das and Brian L Edlow and Daniel C Alexander and Polina Golland and Bruce Fischl},
   city = {United States},
   doi = {10.1126/sciadv.add3607},
   edition = {2023/02/01},
   issn = {2375-2548},
   issue = {5},
   journal = {Science advances},
   month = {2},
   pages = {eadd3607-eadd3607},
   pmid = {36724222},
   title = {SynthSR: A public AI tool to turn heterogeneous clinical brain scans into high-resolution T1-weighted images for 3D morphometry},
   volume = {9},
   url = {https://pubmed.ncbi.nlm.nih.gov/36724222 https://www.ncbi.nlm.nih.gov/pmc/articles/PMC9891693/},
   year = {2023}
}

@article{Edlow2013,
   author = {Brian L Edlow and Robin L Haynes and Emi Takahashi and Joshua P Klein and Peter Cummings and Thomas Benner and David M Greer and Steven M Greenberg and Ona Wu and Hannah C Kinney and Rebecca D Folkerth},
   city = {England},
   doi = {10.1097/NEN.0b013e3182945bf6},
   issn = {1554-6578},
   issue = {6},
   journal = {Journal of neuropathology and experimental neurology},
   month = {6},
   pages = {505-523},
   pmid = {23656993},
   title = {Disconnection of the ascending arousal system in traumatic coma},
   volume = {72},
   url = {https://pubmed.ncbi.nlm.nih.gov/23656993 https://www.ncbi.nlm.nih.gov/pmc/articles/PMC3761353/},
   year = {2013}
}

@article{Billot2023SS,
   author = {Benjamin Billot and Douglas N Greve and Oula Puonti and Axel Thielscher and Koen Van Leemput and Bruce Fischl and Adrian V Dalca and Juan Eugenio Iglesias and ADNI},
   city = {Netherlands},
   doi = {10.1016/j.media.2023.102789},
   edition = {2023/02/25},
   issn = {1361-8423},
   journal = {Medical image analysis},
   month = {5},
   pages = {102789},
   pmid = {36857946},
   title = {SynthSeg: Segmentation of brain MRI scans of any contrast and resolution without retraining},
   volume = {86},
   url = {https://pubmed.ncbi.nlm.nih.gov/36857946 https://www.ncbi.nlm.nih.gov/pmc/articles/PMC10154424/},
   year = {2023}
}

@article{VanEssen2013,
   author = {David C. Van Essen and Stephen M. Smith and Deanna M. Barch and Timothy E.J. Behrens and Essa Yacoub and Kamil Ugurbil},
   doi = {10.1016/j.neuroimage.2013.05.041},
   issn = {10538119},
   journal = {NeuroImage},
   month = {10},
   pages = {62-79},
   title = {The WU-Minn Human Connectome Project: An overview},
   volume = {80},
   year = {2013}
}

@article{Tournier2004,
   author = {J.-Donald Tournier and Fernando Calamante and David G. Gadian and Alan Connelly},
   doi = {10.1016/j.neuroimage.2004.07.037},
   issn = {10538119},
   issue = {3},
   journal = {NeuroImage},
   month = {11},
   pages = {1176-1185},
   title = {Direct estimation of the fiber orientation density function from diffusion-weighted MRI data using spherical deconvolution},
   volume = {23},
   year = {2004}
}

@article{Tuch2002,
   author = {David S. Tuch and Timothy G. Reese and Mette R. Wiegell and Nikos Makris and John W. Belliveau and Van J. Wedeen},
   doi = {10.1002/mrm.10268},
   issn = {0740-3194},
   issue = {4},
   journal = {Magnetic Resonance in Medicine},
   month = {10},
   pages = {577-582},
   title = {High angular resolution diffusion imaging reveals intravoxel white matter fiber heterogeneity},
   volume = {48},
   year = {2002}
}

\newpage
\clearpage

\setcounter{figure}{0}
\setcounter{table}{0}
\renewcommand{\figurename}{Supplementary Figure}
\renewcommand{\tablename}{Supplementary Table}

{\LARGE\bfseries
\noindent Supplementary Materials for: Enhanced Portable Ultra Low-Field Diffusion Tensor Imaging with Bayesian Artifact Correction and Deep Learning-Based Super-Resolution
\par}

\vspace{3em}

\begin{figure}[!htbp]
\centering
\includegraphics[width=0.9\linewidth]{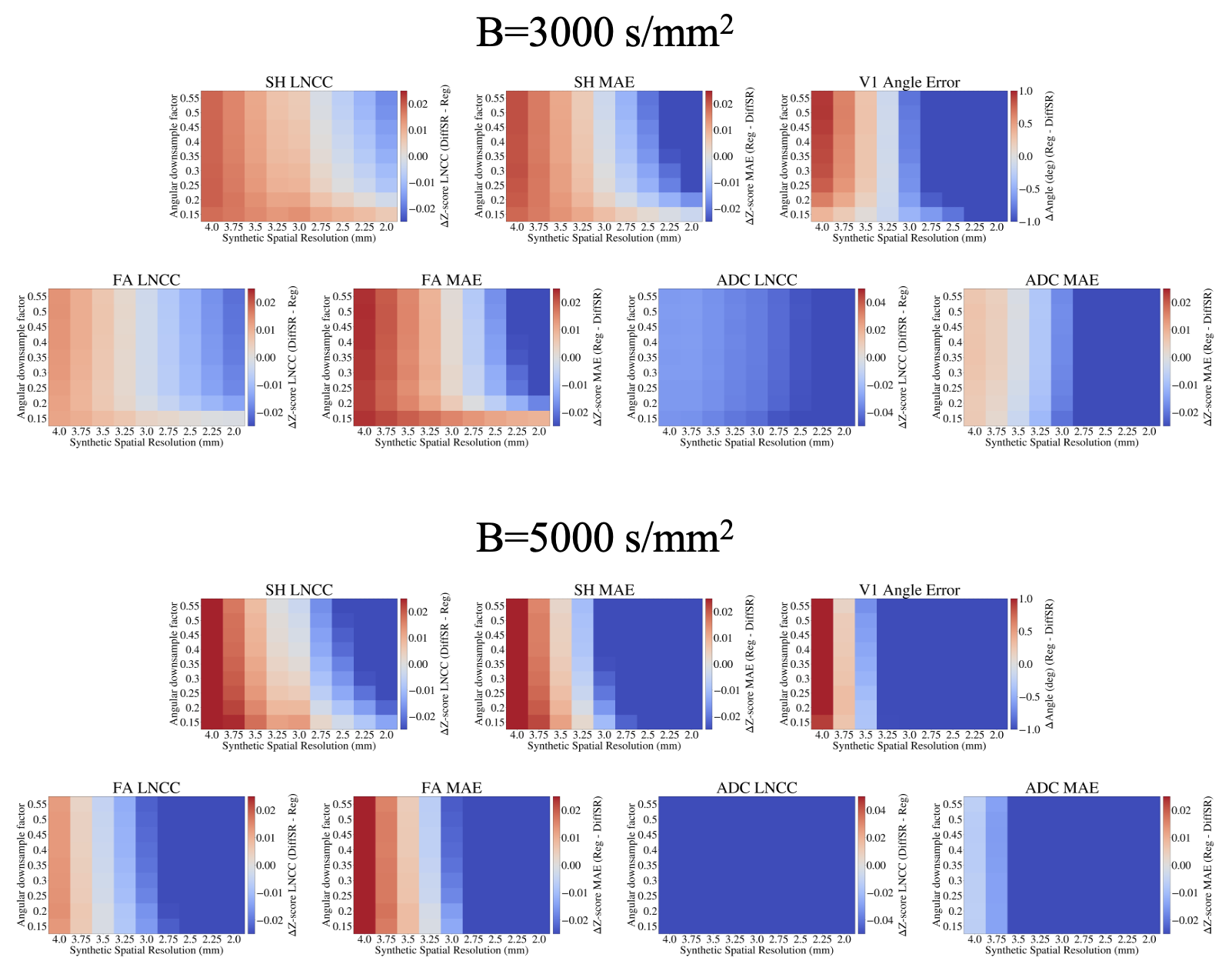}
\caption{\textbf{DiffSR reconstruction accuracy under synthetic spatial and angular downsampling for Connectom HCP data}. DiffSR reconstruction accuracy is shown with respect to standard trilinear upsampling for the b=3000$\frac{s}{mm^2}$ (top) and b=5000$\frac{s}{mm^2}$ (bottom) shells for synthetically downsampled HF DTI data from the MGH HCP dataset. The raw (i.e., gradient directions and low-b’s) HF DTI data were spatially downsampled with trilinear interpolation between 2mm and 4mm isotropic spatial resolutions at 0.25mm intervals. The data was also angularly downsampled by choosing random gradient direction subsets at ratios of 0.15 to 0.55 with respect to the original gradient number in the respective shell. Of note, the b=1000$\frac{s}{mm^2}$ and b=3000$\frac{s}{mm^2}$ shells contain 64 diffusion encoding gradient directions, while the b=5000$\frac{s}{mm^2}$ shell contains 128 diffusion encoding gradient directions, leading to the same gradient subset ratio containing twice as many diffusion encoding gradient directions in the b=5000$\frac{s}{mm^2}$ shell as compared to other shells. Shown are the MAE and LNCC for the SH coefficient channels (channel-wise average), FA and ADC reconstructions. Also shown is the voxel-wise mean angular error for the V1 reconstructions.}
\label{fig:suppF1}
\end{figure}

\newpage

\begin{figure}[!htbp]
\vspace*{10em}
\centering
\includegraphics[width=0.9\linewidth]{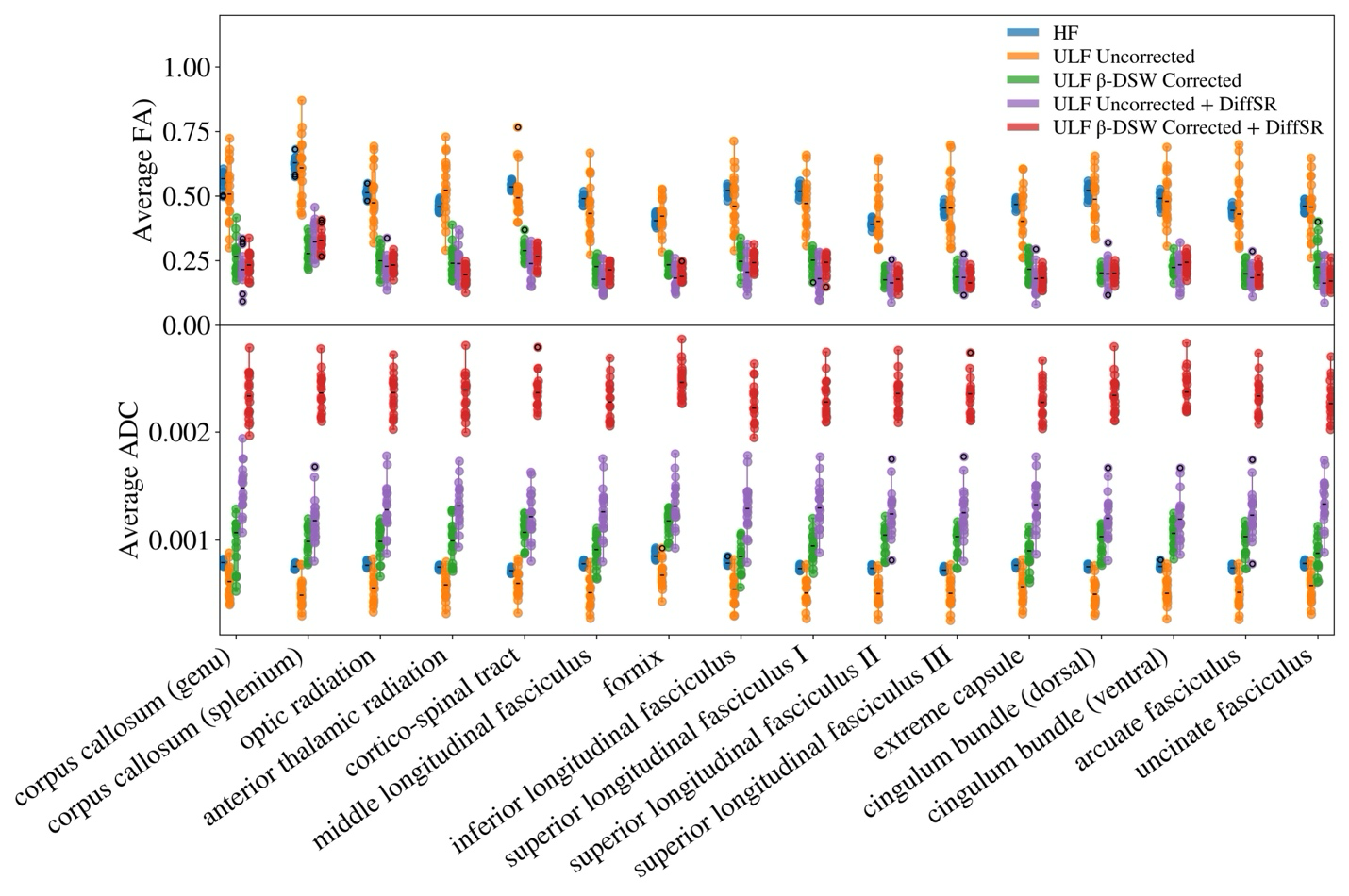}
\caption{\textbf{Per-tract unnormalized fractional anisotropy and apparent diffusion coefficient measurements across ULF DTI reconstruction variants}. Shown (A) are non-z-scored (i.e., un-normalized) distributions of tract-averaged FA (top) and ADC (bottom) for our 18-subject cohort with matched conventional HF DTI and ULF DTI sequences across a subset of white matter tracts segmented with \textit{Tracula}. The native ULF DTI sequence with standard preprocessing (orange) is compared with Beta-DSW bias-corrected ULF DTI without (green) and with superresolution with DiffSR (purple) in terms of overall agreement with matched HF DTI measurements (blue).}
\label{fig:suppF2}
\end{figure}

\newpage

\begin{table}[!htbp]
\vspace*{10em}
\centering
\includegraphics[width=1.5\textwidth, trim=-70 400 0 30, clip]{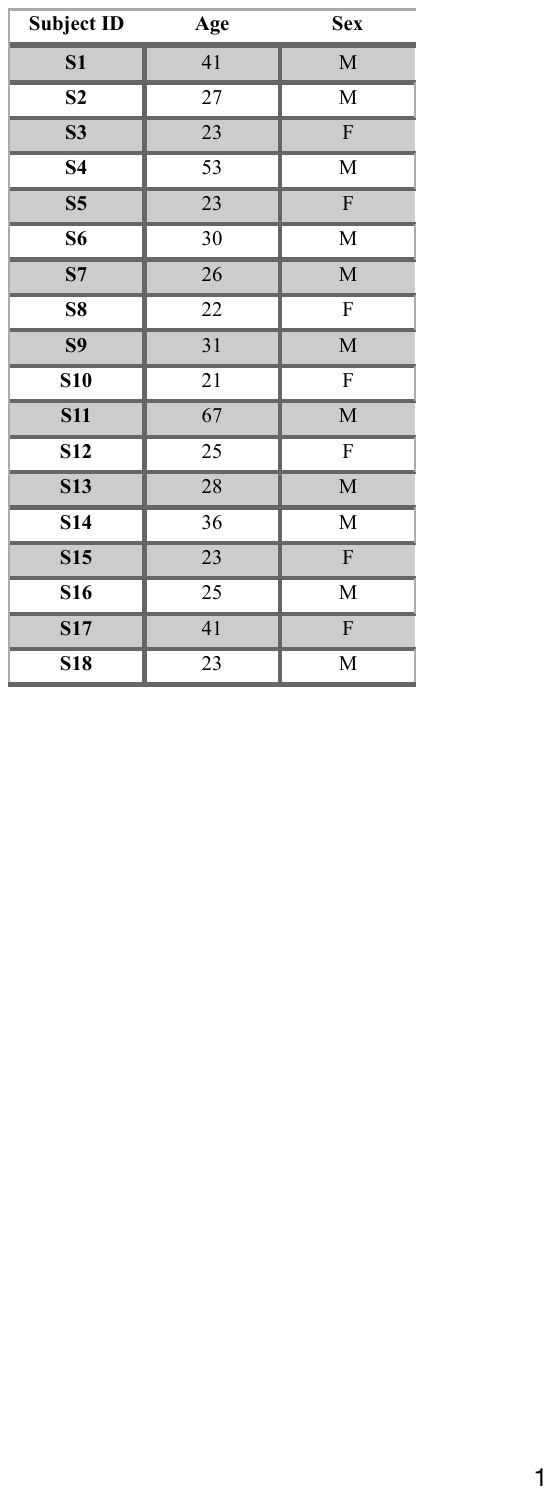}
\caption{\textbf{Subject information from the ULF DTI-matched HF DTI dataset}.}
\label{tab:suppT1}
\end{table}

\begin{table}[!htbp]
\centering
\includegraphics[width=1.2\textwidth, trim=10 180 0 60, clip]{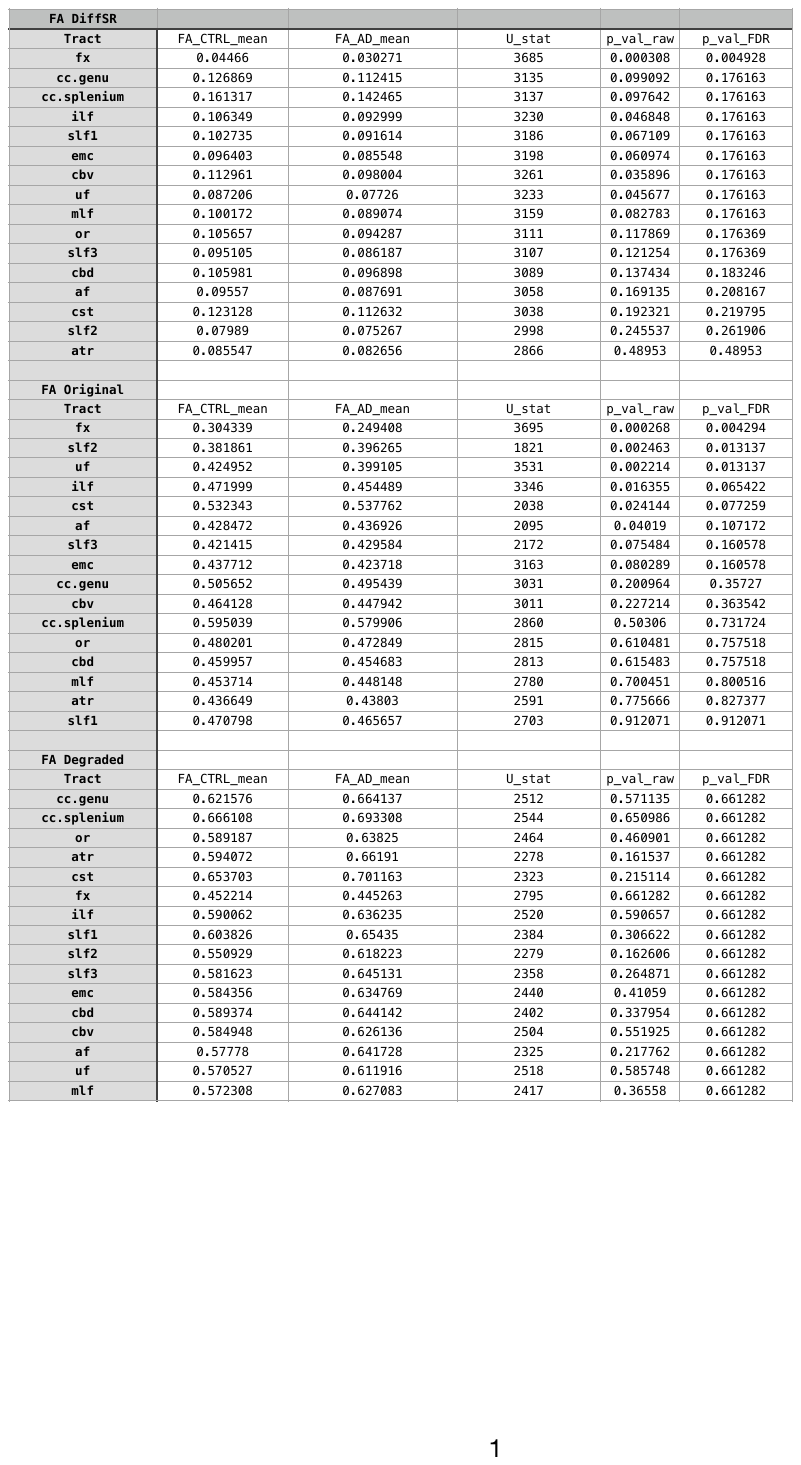}
\caption{\textbf{Statistical information for tract-wise FA differences between control and AD/LMCI ADNI subject groups}. "FA\_CTRL\_mean" and "FA\_AD\_mean" correspond to the tract-wise FA means for the control and AD/LMCI subject groups respectively. “p\_val\_raw” corresponds to the uncorrected two-tailed Wilcoxon rank sum p-value. “p\_val\_FDR” corresponds to the two-tailed Wilcoxon rank sum p-value corrected with Benjamini Hochberg false discovery rate correction. }
\label{tab:suppT2}
\end{table}

\newpage

\begin{table}[!htbp]
\centering
\includegraphics[width=1.2\textwidth, trim=10 180 0 60, clip]{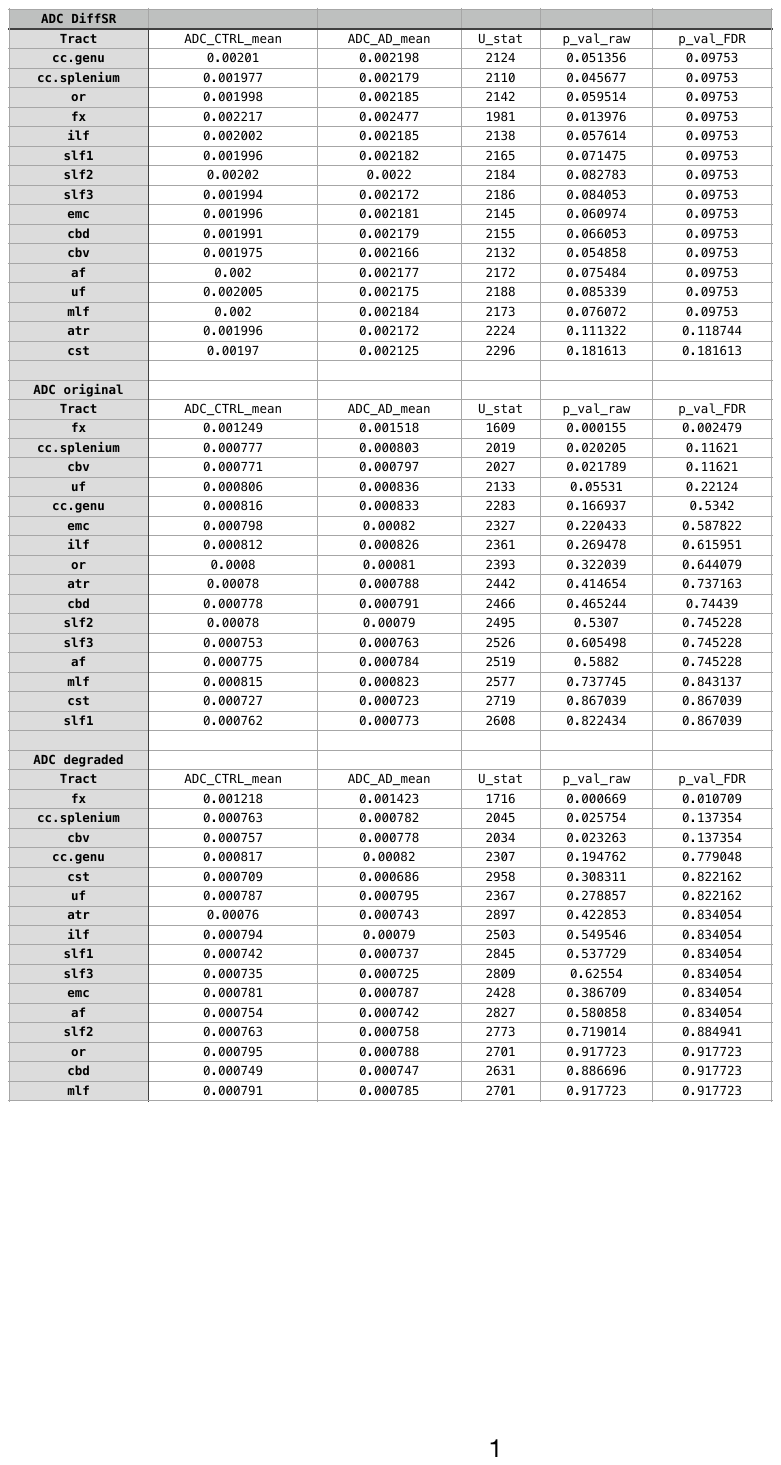}
\caption{\textbf{Statistical information for tract-wise ADC differences between control and AD/LMCI ADNI subject groups}. "ADC\_CTRL\_mean" and "ADC\_AD\_mean" correspond to the tract-wise ADC means for the control and AD/LMCI subject groups respectively. “p\_val\_raw” corresponds to the uncorrected two-tailed Wilcoxon rank sum p-value. “p\_val\_FDR” corresponds to the two-tailed Wilcoxon rank sum p-value corrected with Benjamini Hochberg false discovery rate correction.}
\label{tab:suppT3}
\end{table}

\end{document}